\documentclass[letterpaper]{article}
\usepackage{uai2019}
\usepackage[margin=1in]{geometry}

\usepackage{times}

\usepackage{color}
\usepackage{amsbsy}
\usepackage{amsthm}
\usepackage{amssymb}
\usepackage{amsmath}
\usepackage{setspace}
\usepackage{graphicx}
\usepackage{tikz}
\usepackage{hyperref}
\usetikzlibrary{bayesnet, shapes.symbols, patterns}
\usepackage{multicol}
\newcommand{\mygrid}{\tikz{\draw[step=5mm, thin, dotted] (0,0)  grid (2.4,2.4);}}

\DeclareMathOperator*{\argmin}{arg\,min}
\DeclareMathOperator*{\bel}{Bel}

\usepackage[authoryear]{natbib}

\title{Real-Time Robotic Search using Hierarchical Spatial Point Processes}

\author{} 

\author{ {\bf Olov Andersson} \\
Dept of Computer and Info Science\\
Link\"{o}ping University\\
\texttt{olov.andersson@liu.se} \\
\And
{\bf Per Sid\'{e}n}  \\
Dept of Computer and Info Science\\
Link\"{o}ping University\\
\texttt{per.siden@liu.se} \\
\And
{\bf Johan Dahlin}   \\
Kotte Consulting AB \\
\texttt{work@johandahlin.com} \\
\AND
{\bf Patrick Doherty}   \\
Dept of Computer and Info Science\\
Link\"{o}ping University\\
\texttt{patrick.doherty@liu.se} \\
\And
{\bf Mattias Villani}   \\
Dept of Computer and Info Science\\
Link\"{o}ping University and\\
Department of Statistics\\
Stockholm University\\
\texttt{mattias.villani@gmail.com} \\
}

\begin{document}

\maketitle

\begin{abstract}
Aerial robots hold great potential for aiding Search and Rescue (SAR) efforts over large areas. Traditional approaches typically searches an area exhaustively, thereby ignoring that the density of victims varies based on predictable factors, such as the terrain, population density and the type of disaster. We present a probabilistic model to automate SAR planning, with explicit minimization of the expected time to discovery. The proposed model is a hierarchical spatial point process with three interacting spatial fields for i) the point patterns of persons in the area, ii) the probability of detecting persons and iii) the probability of injury. This structure allows inclusion of informative priors from e.g. geographic or cell phone traffic data, while falling back to latent Gaussian processes when priors are missing or inaccurate. To solve this problem in real-time, we propose a combination of fast approximate inference using Integrated Nested Laplace Approximation (INLA), and a novel Monte Carlo tree search tailored to the problem. Experiments using data simulated from real world GIS maps show that the framework outperforms traditional search strategies, and finds up to ten times more injured in the crucial first hours.\looseness=-1000 
\end{abstract}

\section{Introduction} 
There is rising interest in employing robots such as unmanned aerial vehicles (UAVs) for search and rescue operations. Locating injured victims early is crucial to reduce suffering and mortality rates, and UAVs have the potential to search large areas quickly. Emergency services have begun experimenting with incorporating remote controlled drones or aerial robots with some level of autonomy. However, how to do this efficiently remains an open problem.

Rescue robots is an active research area in AI and robotics, tackling a range of problems from the logistics of aid delivery and communications deployment, down to sensor and motion planning of individual robots. 
Here we focus exclusively on how to improve the search part of the problem, envisioning an automatic and near-optimal solution for aerial search that can be used both as a component of a larger robotic system, or decision support for first-responders. 

Most work on directed exploration in robotics stems from mapping or monitoring of the environment, e.g. \cite{singh2009efficient}, which ignores the full structure of the SAR problem. Even works explicitly on SAR tend to take a narrow view of the problem, such as controlling a sensor to maximize information gain. Heuristics \citep{waharte2010supporting} and exhaustive search via terrain coverage maximization  \citep{huang2001optimal,xu2011optimal} seem common in practice. However, many disasters such as earthquakes, flooding, and even terrorist attacks have a multitude of victims and can cover large areas, where approaches needing human guidance take valuable time away from rescuers. We instead seek a more hands-off probabilistic solution, where humans encode their domain knowledge via priors, and are then free to focus on rescue efforts while the aerial robots do what they do best - scout large areas.\looseness=-1000 

Ideally, the system should be autonomous, robust to the complexities and uncertainty of disasters, and flexible to the requirements of disaster management. Here we leverage probabilistic generative models that reflect the structure of the problem, and solve it as accurately as possible within the stipulated real-time requirements by using approximate probabilistic inference and planning. We need a model that reflects the desiderata that goes into human decision making, and update it in real-time. We propose to separately model the population density, injury probability and detection probability. In conjunction with a terrain-based exploration time, inferring these factors allows us to optimize the sequence of exploratory moves to minimize discovery time. Further, we both allow strong priors on these variables, and can revert back to a spatial process when the priors are uninformative or inaccurate. For example, we can automatically pull estimates of population density from geographic information systems (GIS). A priori, a densely populated area is likely to contain more victims than a sparsely populated one, and a field or road is both faster and easier to explore than a forest. In the case of an earthquake, areas near buildings are likely to contain more victims than roads. During a terrorist attack, whatever early information exists can be encoded and explored first. Without further information, the system should sample high population areas, and if it stumbles on victims, \textit{learn} a local adjustment in the injury model and focus on that area. Although simpler intensity maps and Bayesian methods have previously been used for prioritized search (c.f. \cite{waharte2010supporting,morere2017sequential}), to the best of our knowledge this is the first attempt at capturing the full structure of the problem in probabilistic model that can be updated in real-time. 

However, real-time search in such spatial problems is very challenging for three main reasons: i) only a small part of the point pattern is observed early in the search, ii) the parameter space is high-dimensional and real-time sequential inference is therefore computationally challenging, and iii) optimal search requires solving a computationally hard planning problem under uncertainty, and we need to solve it in real-time.  

The main contributions of this paper are:
\vspace{-1em}
\begin{itemize}
    \setlength\itemsep{-0.2em}
    \item A powerful structured probabilistic model for the search and rescue domain with the possibility to make effective use of prior information 
    \item Real-time inference via deterministic Integrated Nested Laplace Approximation (INLA), so the model can be updated online
    \item A real-time approximation to the search planning problem using a novel variant of Monte-Carlo tree search with options.
\end{itemize} 
The remainder of this paper is organized as follows. We first define the proposed structured probabilistic model and the hierarchical spatial point processes underpinning it. Then we describe how INLA was used for real-time approximate inference in section~\ref{sec:inla}. The search planning problem and the Monte-Carlo tree search approximation is introduced in section~\ref{sec:mcts}, and finally we present results on a range of search and rescue scenarios in section~\ref{sec:experiments}.

\section{A Hierarchical Spatial Point Process for Search-and-Rescue Applications}\label{sec:spatial_processes}

\subsection{Model overview}
Learning a spatial process for real-time search is a difficult inference since only a small part of the point pattern is observed at any given point in time.  We therefore develop a structured hierarchical spatial point process model that allows us to complement the observed point pattern with prior knowledge, for example about the terrain or cell phone traffic in the search area.

Let $\mathbf{Y}=(\mathbf{y}_{1},\ldots,\mathbf{y}_{n})$
denote a spatial \emph{point pattern} over a region of interest $S \subset\mathbb{R}^{2}$, for example the observed spatial locations of $n$ individuals. The simplest example of a \emph{point process} model for such data is the homogeneous Poisson process for which points are uniformly distributed over $S$ with constant intensity $\lambda$.

Search-and-rescue scenarios have a more complex \emph{marked point pattern} where a given person i) may or may not be detected by a searching UAV, and ii) may or may not be injured. We propose a model built up by three interacting spatial fields: 
\begin{itemize}
    \setlength\itemsep{-0.2em}
    \item the population intensity $\lambda(\mathbf{s})$,
    \item  the detection probability $r(\mathbf{s})$,
    \item the probability of being injured $q(\mathbf{s})$,
\end{itemize}
where $\mathbf{s} \in S$. The following subsection gives the details for each of the fields, and how specific prior information can be used in each of the three parts of the process.

\begin{figure*}[ht]
\vspace{1in}
\includegraphics[width=81mm]{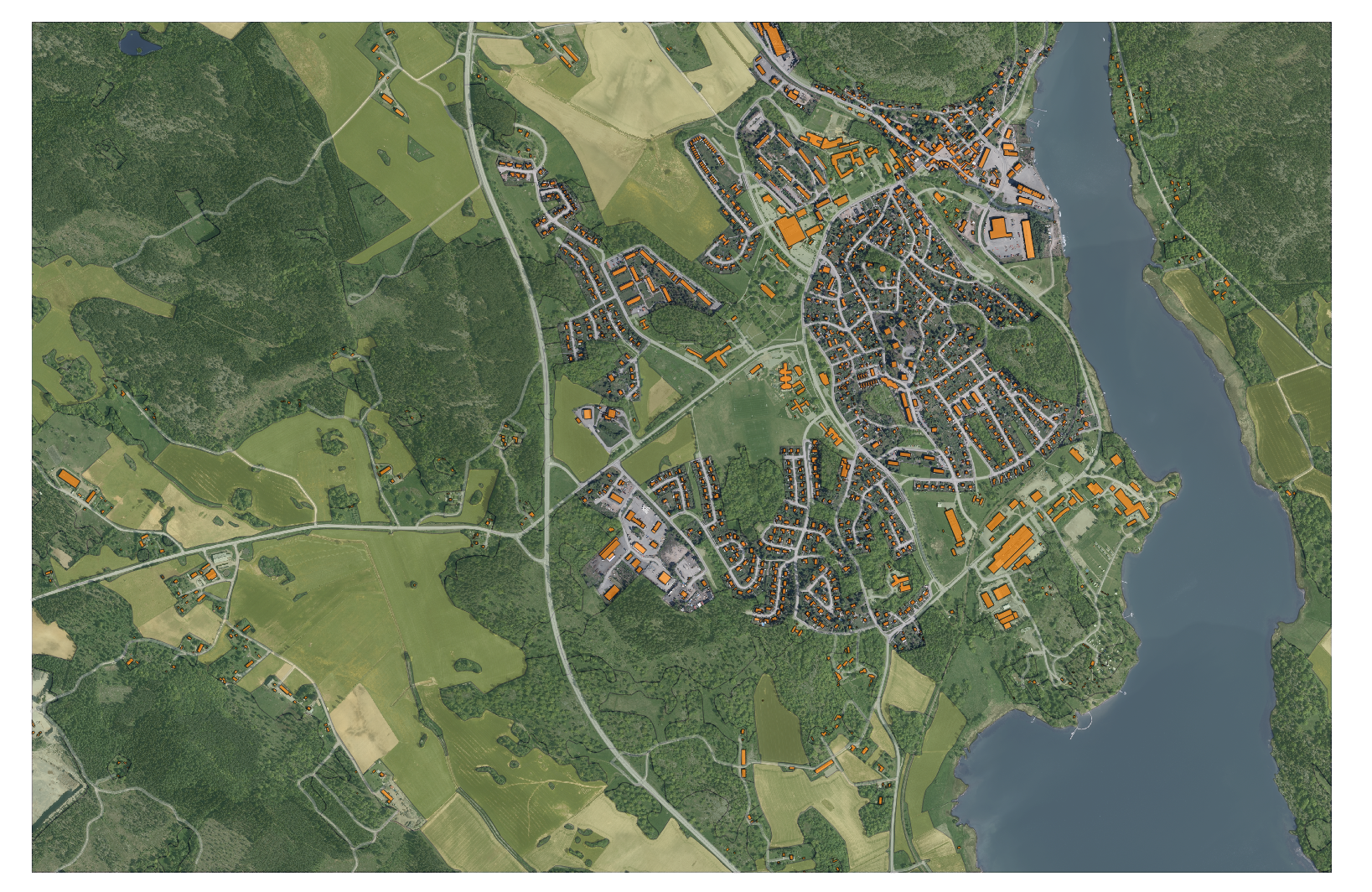}\hspace{0.5cm}
\includegraphics[trim=0mm 3mm 0mm 0mm,width=77mm]{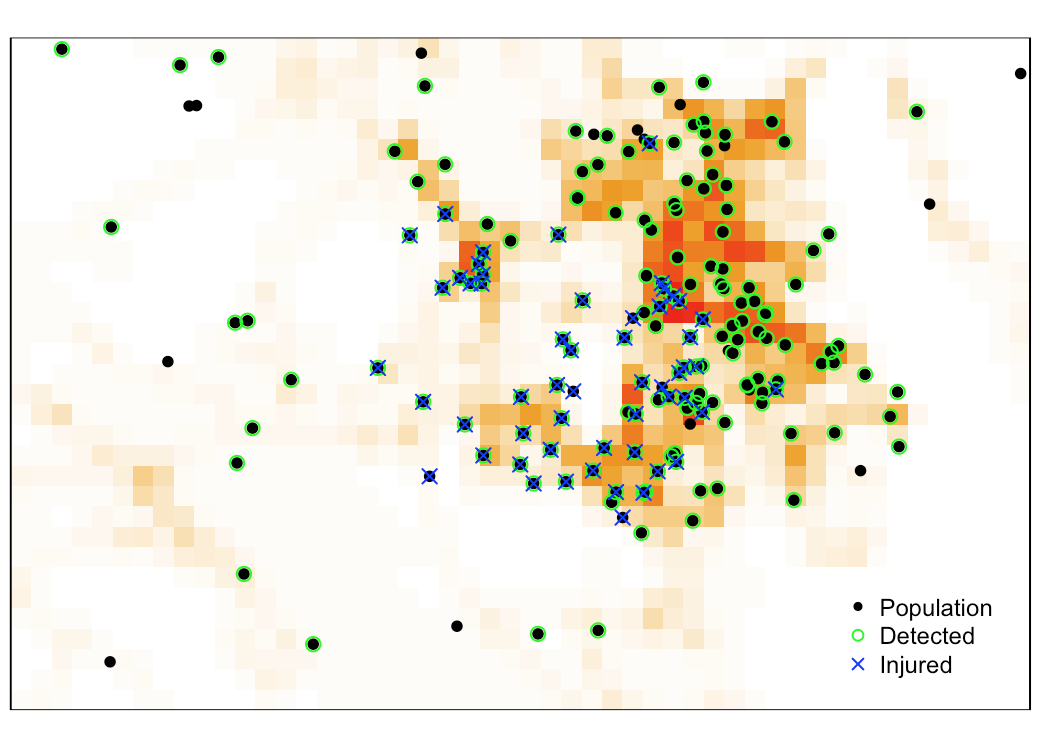}
\caption{Left: Map of the town of Gamleby, Sweden with buildings (orange), forest (dark green), fields (light green), roads (light grey) and water (dark grey) marked out. Right: a sample realization from the model showing the population intensity overlayed by persons in the area (filled dots), detected persons (green circles) and injured persons (blue crosses). There is increased population intensity at buildings and roads, decreased population density in water, lower detection probability in the forest and increased probability of injury due to an explosion in the southwest part of the town. The displayed points have been thinned out by a factor 10 for visualization purposes.}\label{fig:modelRealization}
\end{figure*}

\subsection{Population model}
Let $\mathbf{Y^\star}=(\mathbf{y}_{1}^\star,\ldots,\mathbf{y}_{n}^\star)$ denote the point pattern of persons over a region $S$, and let $N_{y^\star}(\tilde{S})$ denote the number of persons in the subset $\tilde{S} \subset S$. We model the point pattern $\mathbf{Y^\star}$ by a Log Gaussian Cox Process (LGCP, \cite{moller1998log})
\[
N_{y^\star}(\tilde{S})|\lambda\sim\mathsf{Poisson}\left(\int_{\mathbf{s}\in\tilde{S}}\lambda(\mathbf{s})d\mathbf{s}\right),
\]
with log intensity surface given by
\[
\log\lambda(\mathbf{s})=\alpha_{\lambda}+\mathbf{x}_{\lambda}^{\top}(\mathbf{s})\boldsymbol{\beta}_{\lambda}+\xi_{\lambda}(\mathbf{s}),
\]
where $\alpha_{\lambda}$ is an intercept, $\mathbf{x}_{\lambda}(\mathbf{s})$
are spatial covariates, $\boldsymbol{\beta}_{\lambda}$ are regression
coefficients and $\xi_{\lambda}(\mathbf{s})$ is a zero mean Gaussian
Process (GP) over $S$. The spatial covariates $\mathbf{x}_{\lambda}(\mathbf{s})$
could contain any spatial prior information that helps explain the population intensity $\lambda$, for example the location of buildings and water; such information is readily available from GIS systems. The remaining part
of $\lambda$ is modeled as a GP $\xi_{\lambda}(\mathbf{s})$ with a smooth kernel, as people tend to cluster together. We will throughout this
paper focus on GP kernels from the Mat\'ern family.

\subsection{Detection model}
The interpretation of $\mathbf{Y^\star}$ are all persons that could be possibly observed by a UAV if flown over at low height and at good sighting conditions. In practice, conditions may not be perfect and we model the persons actually observed $\mathbf{Y}$ through the detection probability $r(\mathbf{s})$ of observing a person at point $\mathbf{s}$, generating a thinned Poisson process
\[
N_{y}(\tilde{S})|r,\lambda\sim\mathsf{Poisson}\left(\int_{\mathbf{s}\in\tilde{S}}r(\mathbf{s})\lambda(\mathbf{s})d\mathbf{s}\right),
\]
where $N_{y}(\tilde{S})$ denotes the number of observed
persons in $\tilde{S}$. The detection probability $r(\mathbf{s})$ is modeled with
\[
\log r(\mathbf{s})=\mathbf{x}_{r}^{\top}(\mathbf{s})\boldsymbol{\beta}_{r},
\]
where $\mathbf{x}_{r}(\mathbf{s})$ contain prior information about for example terrain type that might affect visibility. $r(\mathbf{s})$ is technically a detection \emph{rate}, since positive values of $\mathbf{x}_{r}^{\top}(\mathbf{s})\boldsymbol{\beta}_{r}$ leads to values for $r(\mathbf{s})$ greater than $1$. However, it can be interpreted as a detection probability when $\mathbf{x}_{r}^{\top}(\mathbf{s})\boldsymbol{\beta}_{r}$ is non-negative for all $\mathbf{s}$, which, if not already the case, could always be achieved by re-balancing the model and increasing the base population level $\alpha_\lambda$ and modifying $\mathbf{x}_{r}^{\top}(\mathbf{s})$ and $\boldsymbol{\beta}_{r}$.

\subsection{Injury model}
We assume that when the disaster strikes all persons have a spatially varying probability $q(\mathbf{s})$ of being injured. Let $\mathbf{w}$ be a binary vector of length $n$ with $w_{i}=1$ iff the $i$th detected person is injured, and $w_{i}=0$ otherwise. This is an example of marked point pattern where each observed point is marked by a binary variable. We assume a geostatistical marking process where the marking process (injured) is independent of the point pattern process (pattern of detected persons) and assumed to follow
\[
w_{i}|q \sim \mathsf{Bernoulli}\left(q(\mathbf{y}_{i})\right),
\]
where
\[
 \log\left(\frac{q(\mathbf{s})}{1-q(\mathbf{s})}\right)=\alpha_{q}+\mathbf{x}_{q}^{\top}(\mathbf{s})\boldsymbol{\beta}_{q}+\xi_{q}(\mathbf{s}),
\]
where $\alpha_{q}$ is an intercept, $\mathbf{x}_{q}(\mathbf{s})$
are spatial covariates, $\boldsymbol{\beta}_{q}$ are regression coefficients
and $\xi_{q}(\mathbf{s})$ is a GP. The covariates $\mathbf{x}_{q}(\mathbf{s})$ could for example be large near buildings in an earthquake scenario.

Figure \ref{fig:modelRealization} displays a sample realization from the model over a map of the town of Gamleby in Sweden. A graphical representation of the model is given later in Figure \ref{fig:graphicalModel} when the inference of the model parameters is discussed.

\section{Real-time online learning using INLA}\label{sec:inla}
The Integrated Nested Laplace Approximation (INLA, \cite{rue2009approximate}) is a fast and memory efficient approximate Bayesian learning algorithm which we show can be successfully applied to our model for sequential learning under real-time constraints.

\subsection{Lattice formulation}\label{sec:inla_lattice}
When learning the model, we assume that the domain $\left(0,x_{1,max}\right)\times\left(0,x_{2,max}\right)$
is rectangular and has been split up into $a_{1}\times a_{2}$ equally
sized rectangles, each with area $\Delta=\frac{x_{1,max}x_{2,max}}{a_{1}a_{2}}$. This approach was previously used for spatial point process models with INLA by \cite{illian2012toolbox}.
Define $n_{ij}$ as the number of detectable persons in cell $s_{ij}$
and $m_{ij}$ as the number of detectable injured persons in $s_{ij}$.
Define $E_{ij}=\Delta$ if $s_{ij}$ has not been visited by the UAV and
$E_{ij}=0$ otherwise. Now, conditional on the latent fields, the joint distribution of the number of persons and injured in cell $s_{ij}$ is given by
\[
n_{ij}|z_{\lambda},z_{r}\sim\mathsf{Poisson}\left(E_{ij}\exp\left(z_{\lambda,ij}+z_{r,ij}\right)\right)
\]
\[
m_{ij}|n_{ij},z_{q}\sim\mathsf{Binomial}\left(n_{ij},\frac{\exp\left(z_{q,ij}\right)}{1+\exp\left(z_{q,ij}\right)}\right)
\]
where $z_{\lambda,ij}$, $z_{r,ij}$, and $z_{q,ij}$ are representative
values of $Z_{\lambda}(s)\equiv \log \lambda(\mathbf{s})$, $Z_{r}(s) \equiv \log r(\mathbf{s}) $ and $Z_{q}(s)\equiv \log(q(\mathbf{s})/(1-q(\mathbf{s})))$ in $s_{ij}$.
Inference is simplified by observing since $\alpha_{q,ij}$ and $\boldsymbol{\beta}_{q}$ only enter the binomial injury model and the log link in both the population and the detection models imply that
\[
z_{\lambda,ij}+z_{r,ij}=\alpha_{\lambda,ij}+\mathbf{x}^\top_{\lambda,ij}\boldsymbol{\beta}_{\lambda}+\mathbf{x}^\top_{r,ij}\boldsymbol{\beta}_{r}+\xi_{\lambda,ij}.
\]
To avoid identification problems between $\boldsymbol{\beta}_{\lambda}$ and $\boldsymbol{\beta}_{r}$ we need to use different covariates in the population process $\mathbf{x}_{\lambda}$ and in the detection process $\mathbf{x}_{r}$. 

\begin{figure}[ht]
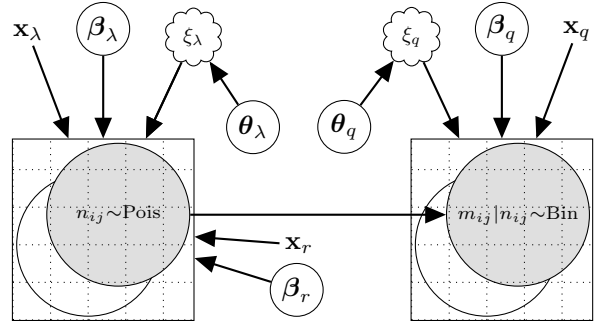

\tikz{
    \node[latent, minimum size=1.9cm, xshift=+0.0cm, yshift = -0.2cm] (nUnObs) {};
    \node[obs, minimum size=1.9cm, xshift=+0.4cm, yshift = +0.2cm] (n) {${\scriptstyle n_{ij}\sim \mathrm{Pois}}$};
    
    \node[rectangle, draw, inner sep=0, xshift=+0.2cm] (poisLattice)
    {
        \mygrid
    };
    \node[cloud, scale=0.8, draw, above=of poisLattice, xshift=1.5cm, yshift=-0.05cm] (xiLambda) {$\xi_\lambda$};
    \node[latent, right=of xiLambda, xshift=-0.9cm, yshift=-1.2cm] (thetaLambda) {$\boldsymbol{\theta}_\lambda$};
    \edge[thick] {thetaLambda} {xiLambda} ;
    \node[const, above=of poisLattice, xshift=-1cm, yshift=0.3cm] (Xlambda) {$\mathbf{x}_\lambda$} ; 
    \node[latent, above=of poisLattice, xshift=0.0cm, yshift=0.1cm] (betaLambda) {$\boldsymbol{\beta}_\lambda$} ; 
    \edge[thick] {betaLambda,xiLambda} {poisLattice} ; 
    \edge[thick, shorten <= 2pt] {Xlambda,xiLambda} {poisLattice} ;

    \node[const, right=of poisLattice, xshift=0.2cm, yshift=-0.2cm] (Xr) {$\mathbf{x}_r$} ; 
    \node[latent, right=of poisLattice, xshift=0.0cm, yshift=-0.8cm] (betar) {$\boldsymbol{\beta}_r$} ;
    \edge[thick] {betar} {poisLattice}
    \edge[thick, shorten <= 2pt] {Xr} {poisLattice} ; 
    
    
    \node[latent, minimum size=1.9cm, xshift=+5.3cm, yshift = -0.2cm] (mUnObs) {};
    \node[obs, right=of n, minimum size=1.9cm, xshift=+2.4cm] (m) {${\scriptstyle m_{ij}\vert n_{ij} \sim \mathrm{Bin}}$};
    \edge[thick] {n} {m} ;
    \node[rectangle, draw, inner sep=0, xshift=+5.5cm] (binLattice)
    {
        \mygrid
    };
    \node[cloud, scale=0.8, draw, above=of binLattice, xshift=-1.5cm, yshift=-0.05cm] (xiq) {$\xi_q$};
    \node[latent, left=of xiq, xshift=+0.8cm, yshift=-1.2cm] (thetaq) {$\boldsymbol{\theta}_q$};
    \edge[thick] {thetaq} {xiq} ;
    \node[const, above=of binLattice, xshift=1cm, yshift=0.3cm] (Xq) {$\mathbf{x}_q$} ; 
    \node[latent, above=of binLattice, xshift=0.0cm, yshift=0.1cm] (betaq) {$\boldsymbol{\beta}_q$} ; 
    \edge[thick] {betaq,xiq} {binLattice} ;
    \edge[thick, shorten <= 1pt] {Xq} {binLattice} ;

}
\caption{Graphical model representation of the lattice formulation of the model at a given search iteration. The chequered squares represent lattices and the nodes inscribed in them represent the variables $n_{ij}$ and $m_{ij}\vert n_{ij}$ over the lattice (grey = observed, white = unobserved). Cloud nodes represent Gaussian Processes over $\mathbb{R}^2$.}\label{fig:graphicalModel}
\end{figure}

A graphical representation of the model is given in Figure \ref{fig:graphicalModel}.

\subsection{Integrated Nested Laplace Approximation}
Inference in models with high-dimensional spatial random fields is a challenging problem, and exact methods such as Markov Chain Monte Carlo (MCMC) are much too slow for real-time learning. Variational inference is the go-to method in machine learning for fast Bayesian inference, but is well known to underestimate posterior uncertainty, particularly in high-dimensional spatial problems, see e.g. \cite{rue2009approximate}.

INLA \citep{rue2009approximate} is a framework for fast accurate approximation of Bayesian posterior distributions in the class of latent Gaussian models. INLA is by now a standard method for spatial problems in the statistical literature, but is rarely used in Robotics. To describe the class of latent Gaussian models, let $\mathbf{z} \in \mathbb{R}^d$ be a (high-dimensional) vector of Gaussian variables with prior $\pi(\mathbf{z} \vert \boldsymbol{\theta})=N(\mathbf{0},Q(\boldsymbol{\theta})^{-1})$, where $Q(\boldsymbol{\theta})$ is a sparse precision (inverse covariance) matrix. In our case $\mathbf{z}$ contains the Gaussian random fields $\xi_{\lambda}(\mathbf{s})$, $\xi_q(\mathbf{s})$ as well as the fixed effects $\beta_\lambda$, $\beta_r$ and $\beta_q$. The (low-dimensional) vector of hyperparameters in the prior $\boldsymbol{\theta}$ contains the variances and length scales of the Mat\'ern kernel functions for $\xi_{\lambda}(\mathbf{s})$ and  $\xi_q(\mathbf{s})$, and unknown hyperparameters in the priors for the fixed effects. INLA further assumes the $n$ observations in $\mathbf{y}$ are independent conditional on the latent variables $\mathbf{z}$, with likelihood function
\begin{equation}\label{eq:INLAlikelihood}
\pi(\mathbf{y} \vert \mathbf{z}, \boldsymbol{\theta}) = \prod_{i=1}^n \pi(y_i \vert z_i, \boldsymbol{\theta}). 
\end{equation}
The vector $\boldsymbol{\theta}$ can include additional hyperparameters needed to describe $\pi(y_i \vert z_i, \boldsymbol{\theta})$. Our likelihood for the Poisson model for detected persons, $n_{ij}$, and the likelihood of the binomial model for number of injured persons, $m_{i,j}$, are clearly both of the form \eqref{eq:INLAlikelihood}.


INLA uses an intricate mix of several Laplace approximations for the high-dimensional $\mathbf{z}$ combined with numerical integration of the low-dimensional hyperparameters $\boldsymbol{\theta}$ to approximate the marginal posteriors of the latent variables $z_i$ and the joint posterior of the hyperparameters $\pi (\boldsymbol{\theta}\vert \mathbf{y})$. The basic INLA approximation is of the form
\[
\pi(z_i \vert \mathbf{y}) \approx \int \tilde \pi(z_i \vert \boldsymbol{\theta},\mathbf{y}) \tilde \pi(\boldsymbol{\theta}\vert\mathbf{y})d\boldsymbol{\theta}.\label{eq:INLAmargApprox}
\]
where $\tilde \pi(z_i \vert \boldsymbol{\theta},\mathbf{y})$ is obtained by marginalizing a Laplace approximation \citep{tierney1986accurate} of $\pi(\mathbf{z} \vert \boldsymbol{\theta},\mathbf{y})$ and 
\[
\tilde \pi(\boldsymbol{\theta}\vert\mathbf{y}) = \frac{\pi( \mathbf{z},\boldsymbol{\theta}, \mathbf{y})}{\tilde \pi_G (\mathbf{z} \vert \boldsymbol{\theta}, \mathbf{y} )} \bigg\vert_{\mathbf{z}=\mathbf{z}^\star(\boldsymbol{\theta})}, 
\]
where $\tilde \pi_G (\mathbf{z} \vert \boldsymbol{\theta}, \mathbf{y} )$ is a Gaussian approximation to the full conditional posterior of $\mathbf{z}$ and $\mathbf{z}^\star$ is the mode of $\mathbf{z}$ for a given $\boldsymbol{\theta}$. The integral in \eqref{eq:INLAmargApprox} is performed numerically by summing over a set of carefully selected support points in $\boldsymbol{\theta}$, see \cite{rue2009approximate} for details. Note that INLA does not approximate $\pi(\boldsymbol{\theta}\vert\mathbf{y})$ by a Gaussian, which is important since $\pi(\boldsymbol{\theta}\vert\mathbf{y})$ is often highly non-Gaussian.

The use of \emph{nested} Laplace approximations makes INLA extremely accurate for latent Gaussian models, see for example \cite{rue2009approximate} and \cite{teng2017bayesian} for some evidence for the Log Gaussian Cox Process. In particular, INLA has been shown to be much more accurate than variational approximations.  

INLA is many orders of magnitude faster than Markov Chain Monte Carlo (MCMC) and Hamiltonian Monte Carlo (HMC), and can therefore be successfully applied in a real-time context. By exploiting the sparsity of the precision matrix $Q(\boldsymbol{\theta})$ that results from conditional independencies that appear naturally in spatial and temporal problems and efficient reordering schemes \citep{rue2005gaussian}, INLA scales favorably as $O(d^{3/2})$ in 2D, where $d$ is the total number of cells where the fields are evaluated. Moreover, since our focus here is on $\pi(x_i \vert \mathbf{y})$ rather than hyperparameter inference per se, we will use the Empirical Bayes (EB) to optimize wrt $\boldsymbol{\theta}$. This gives additional speed-ups since we can also benefit from a warm start with excellent initial values from the previous search iteration, followed by an fast update of the posterior $\pi(x_i \vert \mathbf{y},  \boldsymbol{\hat\theta}_{EB})$.

The core of INLA is implemented in C++ with the convenient \textsf{r-inla} interface to the statistical programming language R, see \cite{rue2017bayesian} for details.

\section{Planning Exploratory Moves via MCTS}\label{sec:mcts}
The search part of search and rescue has traditionally been solved by a variety of exploration methods, as e.g. a coverage maximization problem \citep{huang2001optimal}, travelling salesman problem to minimize a path cost, or as information maximization in POMDPs \citep{waharte2010supporting}. However,  attempting to solve these sequential decision problems optimally usually result in computationally hard problems. Greedy heuristics or simple coverage algorithms, such as lawnmower patterns, are therefore often employed in search and rescue applications. 

As we employ a sophisticated probabilistic model, it is natural to view the decision problem as a partially observable Markov decision process, or POMDP \citep{ASTROM1965174}. Only the visited regions are observed, which we use to infer a belief over the remaining spatial point process.

Formally, we define the search state of our problem to be $\mathbf{x} = (\bel(\mathbf{m}), \mathbf{p})$. The matrix $\mathbf{m}$ consists of all $m_{ij}$, random variables for the number of injured in the $\Delta$-sized discretization of the spatial point process from section~\ref{sec:inla_lattice}. We use the notation $\bel(\mathbf{m})$ for the distribution of $\mathbf{m}$. We also define $\mathbf{p} \in \{s_{ij}:\forall\,(i,j)\}$ as the position of the UAV in the cell grid, which can be  considered known by GPS at the lengthscales of our cell size. As the $\bel(\mathbf{m})$ are probability distributions, this is technically a belief-augmented MDP formulation (c.f. \cite{thrun2005probabilistic}). 

The UAV has to sequentially decide on which cell to explore, in the form of actions $a \in \mathcal{A} = \{s_{ij} : \forall\,(i,j)\}$ that map uniquely to exploration of a cell $s_{ij}$. Cell exploration takes time  $T_{ij}= \mathbf{x}_r^{\top}(\mathbf{s})\boldsymbol{\beta}_{T}$, where $\mathbf{x}_r(\mathbf{s})$ are the aforementioned spatial covariates for terrain type from section~$\ref{sec:inla}$, and $\boldsymbol{\beta}_T$ are user-supplied estimates of their respective exploration time. However, the UAV cannot teleport between cells, it additionally takes time $T_f \cdot \mathrm{dist}(\mathbf{p}, s_{ij})$ to reach a non-adjacent cell $s_{ij}$. Clearly, a sequential exploration of adjacent cells takes less time but may not discover the most injured, resulting in a difficult trade-off. 

Finally, we want to solve the optimal exploration problem
\[
\argmin_{\pi(\mathbf{x})} \; \mathbb{E}_{\boldsymbol{\tau}|\pi(\mathbf{x})}[c(\boldsymbol{\tau})],
\] 
where $c(\boldsymbol{\tau})$ is a cost function and $\boldsymbol{\tau}=\{\mathbf{x}_{t_i}...\mathbf{x}_{t_N}\}$ is the trajectory through the belief-augmented state space $\mathbf{x}_t$, from current time $t_i$ until all cells have been explored at time $t_N$. The trajectory is uncertain as the injuries $\mathbf{m}$ are partially unknown at decision time. By taking action $a_t = \pi(\mathbf{x}_t)$, where $a_t$ maps to a cell $s_{ij}$, data $o_t=(n_{ij},m_{ij})$ will be observed and beliefs will be updated by $\bel_{t+1}(\mathbf{m}) = f(\bel_{t}(\mathbf{m}),a_t,o_t)$, where the transition function $f$  updates the spatial point process as described in section \ref{sec:inla}.  

While maximizing information is popular in POMDP formulations of search (c.f. \cite{morere2017sequential, waharte2010supporting}), not all information is equally useful in more complex models, e.g. one may have large uncertainty but low expectation in sparsely populated areas. Contrary to earlier work, we take a more direct approach that at each time $t_i$ attempts to minimize the total remaining harm to victims,
\[
c(\boldsymbol{\tau}) = \int_{t_i}^{t_N} \sum_{ij} m^*_{ij}(t)h(t)\,dt,
\]
where $m^*_{ij}(t)$ is the number of \textit{unexplored} injured in cell $s_{ij}$ at time $t$, and $h(t)$ is the rate of harm. This is integrated with the trapezoid rule over the varying durations of the discrete sequence of actions. For convenience, in the following we assume the rate of harm (e.g. mortality) is proportional to the time spent undiscovered, $h(t) \propto 1$.

Unfortunately, the complexity of solving POMDPs is doubly-exponential \citep{thrun2005probabilistic}. Even with a discretized state lattice and four directional actions, this problem will be infeasible for real-world sizes. 

Monte-Carlo tree search (MCTS) is an approximate solution to discrete sequential decision making problems. In its purest form, MCTS is a tree search algorithm that treats the problem of finding good branches (actions) as a sequence of bandit problems solved by the UCB (Upper Confidence Bounds) algorithm. The resulting UCT algorithm  \citep{kocsis2006bandit} effectively treats finding good plans as an exploration problem in itself. By guiding the search, the effective branching factor can be significantly reduced.

While most well-known for its successes in the game of Go \citep{browne2012survey,silver2017go}, MCTS has recently found uses in motion planning problems \citep{hennes2015interplanetary}. In  \cite{morere2017sequential} MCTS was also used to plan belief trajectories in a POMDP for environment monitoring. However, POMDP planning is still very expensive, and they could only afford to plan three steps forward. The reason is that for each tested action in the search tree, uncertain outcomes have to be sampled, and new beliefs inferred as in section~\ref{sec:inla}.

To create an effective MCTS algorithm for our search domain we propose three modifications, i) incorporating action-abstractions, a small set of long-range moves $\mathcal{A}_{\mathrm{max}}$ directly to the cells $s_{ij}$ with the maximum expected number of unexplored injured $m^*_{ij}(t_i)$, ii) a certainty equivalence assumption where random variables are replaced with their expected values, and iii) using receding-horizon planning with warm-starts and a fast domain-specific cost-to-go approximation.

\begin{figure}[ht]
\includegraphics[clip,width=\linewidth]{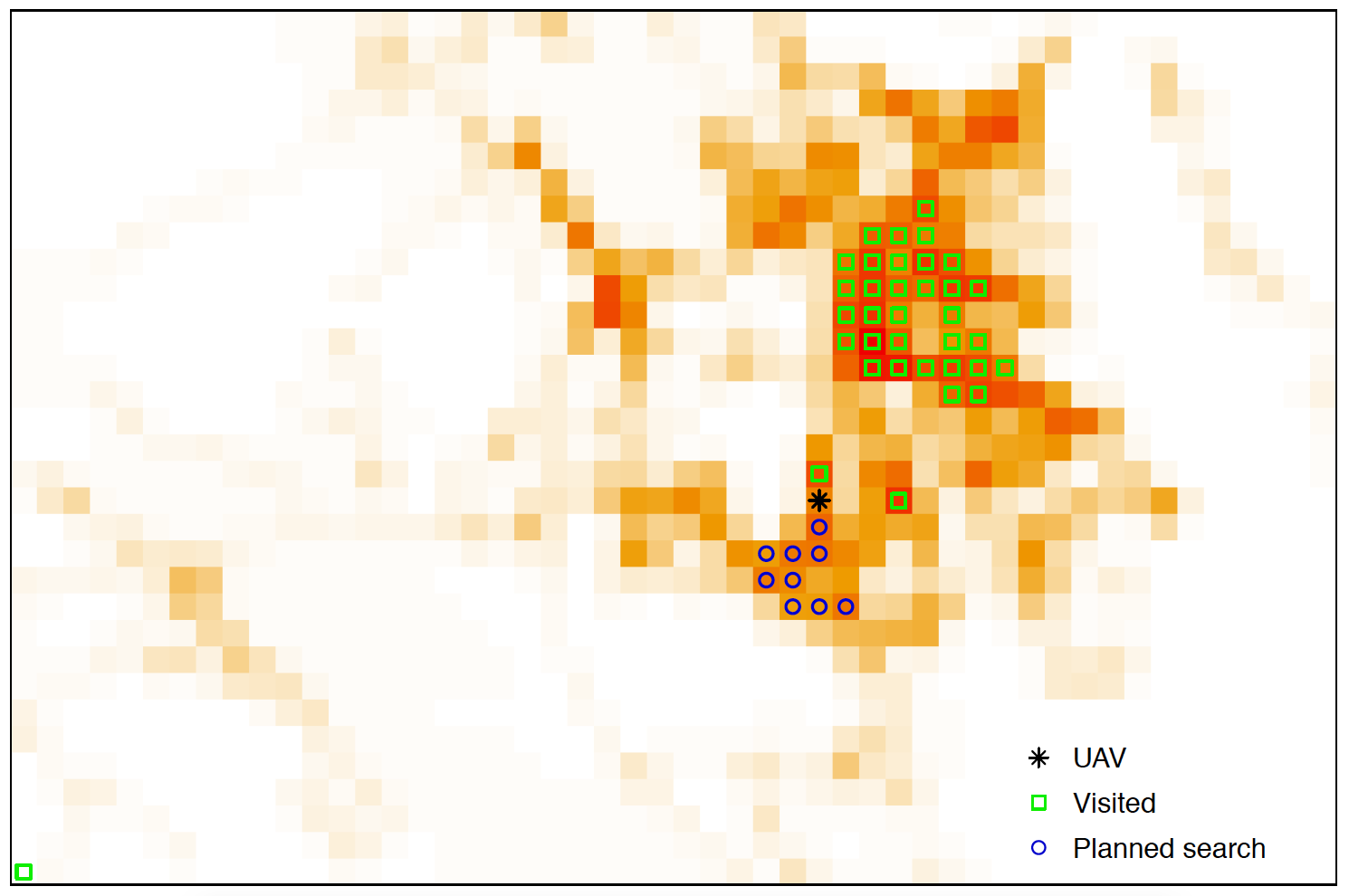}
\caption{\label{fig:MCTS_example}Search scenario using \texttt{MCTSJump}, overlaid on heatmap for expected number of detectable injured.}
\end{figure}

We test two variants. The first, simply called \texttt{MCTS}, can only move through adjacent squares  $\mathcal{A}_\mathtt{MCTS} = \{s_{ij} : \operatorname{adj}(\mathbf{p}, s_{ij})\}$, to explore, or if explored, fly through. The second, called \texttt{MCTSJump}, additionally includes the long-range moves from i) in the first time step $t_i$ of each plan, i.e. $\mathcal{A}_{\mathtt{MCTSJump},t_i} =  \mathcal{A}_{\mathrm{max}}\,\cup\,\mathcal{A}_\mathtt{MCTS}$. An example search pattern from \texttt{MCTSJump} can be seen in Figure~\ref{fig:MCTS_example}.

These action-abstractions can be seen as a type of option policies \citep{stolle2002learning,subramanian2016efficient} in reinforcement learning. Adding more actions increases the branching factor, which results in a difficult trade-off. However, we found that instead of selecting target cells equally spaced or random, using just the top ten best cells resulted in a noticeable performance increase. This also has the benefit of remedying the biggest drawback of iii), the finite planning horizon often made it leave some cells unexplored in the end. 

The certainty equivalence assumption $\bel_{t+1}(\mathbf{m}) \approx f(\delta_{\mathbb{E}_t(\mathbf{m})}(\mathbf{m}),a_t,o_t)$, where $\delta_a(\cdot)$ is the Dirac spike at $a$, allows us to forego sampling from the outcomes to update the belief model, which is the main bottleneck of planning in POMDPs. This alone let us effectively plan $20$ steps ahead. However, it can be a strong assumption, because plans are \textit{evaluated} on the premise that the future is predictable, which means it will not value recourse, the possibility to later change the plan if it turns out worse than expected. In practice it still replans at each step however, and due to the spatial correlation in our problem, we also argue that the effect is small. Significant recourse is costly - it often takes several moves to see large changes. 

Finally, the receding-horizon formulation from iii) is standard in control, where it is sometimes known as model-predictive control. By cutting the planning horizon from $t_N$ to $t_H$ and adding a domain-specific approximation of the remaining cost $c(\boldsymbol{\tau})=c(\boldsymbol{\tau}_{t_i..t_H})+\hat{c}_{t_H..t_N}(\mathbf{x}_{t_H})$, computation cost can be greatly decreased. Warm-starts at from the best plan at $t_{i-1}$ allows computation to be amortized over several iterations. Here we just assume the cost-to-go decreases linearly to zero as for an ideal lawnmower pattern. This allows us to compare plans of different duration, such as action-abstractions moves from i). 

As MCTS is an any-time algorithm we give it a fixed 3 second compute budget. It is implemented in C++ and evaluates about 100 000 plans. As it does not assume any fixed search pattern or observation order, it also allows human operators the flexibility to take control if needed.

\section{Experiments}\label{sec:experiments}

Here we test the proposed real-time probabilistic SAR framework and MCTS exploration algorithm. To the best of our knowledge, this is the first principled attempt at probabilistic modelling of the full complexity of the SAR problem, and explicit minimization of harm, in real-time or otherwise. As noted in the introduction, related probabilistic approaches appear to have a narrower focus in what they model. On the planning side, existing principled optimization-based approaches typically scale poorly  \citep{waharte2010supporting,morere2017sequential} to the large scenarios we envison. Maximum coverage algorithms \citep{xu2011optimal,huang2001optimal}, or some heuristic combination of locally and globally greedy behavior, are often used in practice. Unfortunately, heuristics tend to be sensitive to scenario-specific tuning. We therefore use a maximum-coverage algorithm as baseline. Fortunately, tree-search algorithms like \texttt{MCTS} subsume locally-greedy algorithms if planning horizon is taken to zero, and should in expectation be a dominating strategy as the planning horizon is increased. Further, our proposed \texttt{MCTSJump} algorithm will also always include the globally-greedy choice in its list of action abstractions, and can therefore be seen as a principled way of solving this trade-off as an optimization problem, rather than as a heuristic needing tuning to a scenario. 

We use real-world GIS data from the Swedish government, which offered easy access to a wealth of data on favorable terms\cite{lantmateriet}. We selected a 4.0x2.7km area around the town Gamleby seen in Figure~\ref{fig:modelRealization}. It contains a variety of terrain and is within a proposed UAV test zone, where we may be permitted to test the algorithm with real UAVs. 

Data was simulated using the the hierarchical spatial point process model in section~\ref{sec:spatial_processes}, then discretized to 50x33 lattice cells for search and model inference, each about the size of a soccer field.  We assume the UAV can fly at a speed of $10$m/s for fly-to moves. The cell explore times are set to $\boldsymbol{\beta_T}=\{1,2,0.5,0.5,0.75\}$ minutes for terrain covariates "buildings", "forest", "road", "field", and "water" respectively. This reflects a fast overhead search with some compensation for difficult terrain such as forests needing multiple angles. The detection covariate for forest was similarly set lower. This rapid search pace also highlights the importance of real-time performance, and why we capped MCTS to 3 seconds. For reference, the entire inference and planning loop in our prototype implementation takes about 5 seconds on a Core i7 CPU, imposing minimal overhead on the search. 

In the following we test four scenarios reflecting different types of real-world disasters and the level of prior information available. A summary of the scenarios, the covariates used in the data generating process, as well as those used for the inference models, can be seen in Table~\ref{table:Scenarios}. Each scenario is replicated 15 times from different seeds, except for the first one, which used 30. As our mission area is rectangular, a simple baseline coverage algorithm is a zick-zack, or "lawnmower" pattern. In terms of covering the largest area in the shortest time, this is optimal. In all cases, we attempted to find reasonably wide distributions for parameters in the data generating processes, $\boldsymbol\beta_\lambda$, $\theta_\lambda$, $\boldsymbol\beta_\lambda$ and $\theta_\lambda$, by sampling realizations of the spatial point process and comparing to real-world expectations. We also experimented with perturbing the inference priors and found remarkable small changes in the results, see the supplementary material.

\begin{table*}
\caption{Scenario Settings. Covariates and spatial fields in data generating
process and inferred model. \protect \\
$B$=buildings, $R$=roads, $W$=water, $F$=forest, $G_{i}$=Gaussian no $i$, $S$=spatial field. Deviations from truth in red.}
\label{table:Scenarios}
\centering{}%
\begin{tabular}{llllllll}
\hline 
 &  &  &  & Scenario A & Scenario B & Scenario C &  Scenario D \tabularnewline
 
 &  & Population &  & $B$ & $B+R+W+S$ & $B+R+W+S$ &   $B+R+W+S$\tabularnewline
\textbf{Truth} &  & Detection &  & $-$ & $F$ & $F$ &   $F$\tabularnewline
 &  & Injury &  & $-$ & $B+S$ & $G_{1}$ &   $G_{2}+G_{3}$\tabularnewline
 &  &  &  &  &  &  &   \tabularnewline
 
 &  & Population &  & ${\color{red}S}$ & $B+R+W+S$ & $B+R+W+S$ &   $B+R+W+S$\tabularnewline
\textbf{Model} &  & Detection &  & $-$ & $F$ & $F$ &   $F$\tabularnewline
 &  & Injury &  & $-$ & $B+S$ & ${\color{red}G_{1}+S}$ &   ${\color{red}G_{2}+S}$\tabularnewline
 
\hline 
\end{tabular}
\end{table*}

\begin{table*}
\caption{Time until half of injured have been found}
\label{table:Halftime}
\centering{}%
\begin{tabular}{lrrrrr}
\hline 
 &  & Scenario A & Scenario B & Scenario C & Scenario D \tabularnewline
 Lawnmower &   & 835 &  1113 & 164 & 271 \tabularnewline   
 MCTS      &   & 273 &  118 & 88 & 121 \tabularnewline
 MCTSjump  &   & 241 &  98 & 69 & 100 \tabularnewline  
\hline 
\end{tabular}
\end{table*}


\subsection{Scenario A: No prior information}
In this scenario we assume the model does not have access to any useful spatial covariates. While in practice some information tends to be available, this was designed to test the capability of the model to fall back to the spatial fields, to cover for unexpected situations.

The ground truth is a population distribution drawn from GIS building covariates not available to the agent, see Table~\ref{table:Scenarios}. For simplicity we ignore the injury part of the model and focus only on maximizing the number of people found in this scenario. As can be seen from the results in the top row of Figure~\ref{fig:scenario2prop} and Table~\ref{table:Halftime}, our model with \texttt{MCTS} and \texttt{MCTSJump} significantly outperforms the lawnmower coverage strategy. Just relying on the spatial field was sufficient to capture the natural clustering in population data. In this case however, the improvements offered by long-range moves ("jumps") was not statistically significant, which is not surprising considering the spatial correlation captured by the field only gives local information.

\subsection{Scenario B: Earthquake}
\vspace{-0.7em}
Here we simulate a classical earthquake scenario. We generate population using all five GIS covariates, as well as a spatial field. As this is an earthquake, both building covariates and a field was used to draw realizations of injured people. We use the same structure of the model for inference, and reasonably uninformative priors.
Figure~\ref{fig:scenario2prop} and Table~\ref{table:Halftime} show that by drawing on the GIS covariates, even corrupted by a spatial field, our algorithm is significantly  faster than in A, and also increases its lead to lawnmower by a wide margin. This also showcases the advantage of \texttt{MCTSJump}, which outperformed regular \texttt{MCTS} by first drawing on the injury covariate to make informed jumps directly to the urban areas, then using jumps at the end to complete the map. Finite-horizon MCTS can leave some areas unvisited.

\subsection{Scenario C: Terrorist attack - known site}
\begin{figure*}[h!]
\vspace{1in}
\includegraphics[width=40mm]{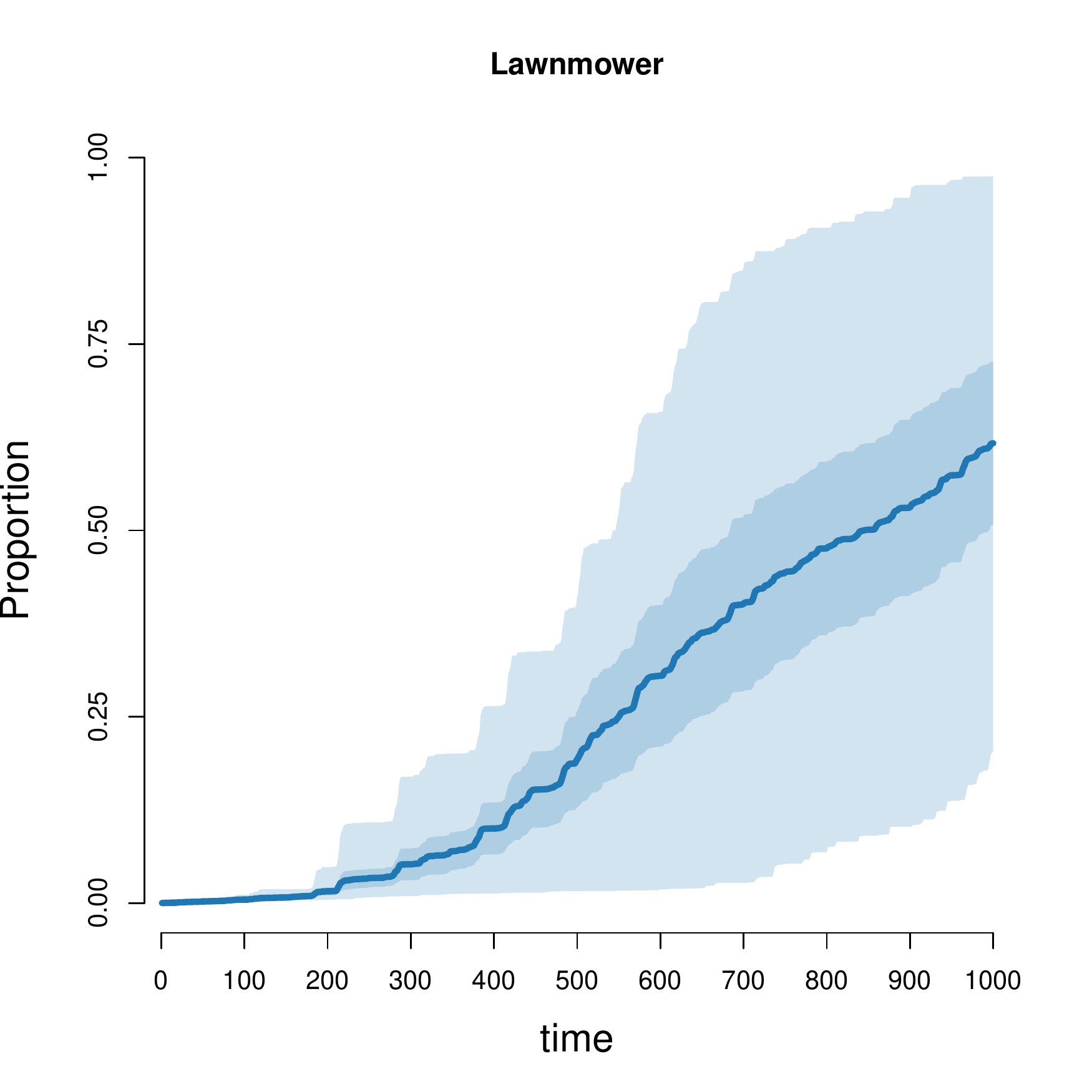}
\includegraphics[width=40mm]{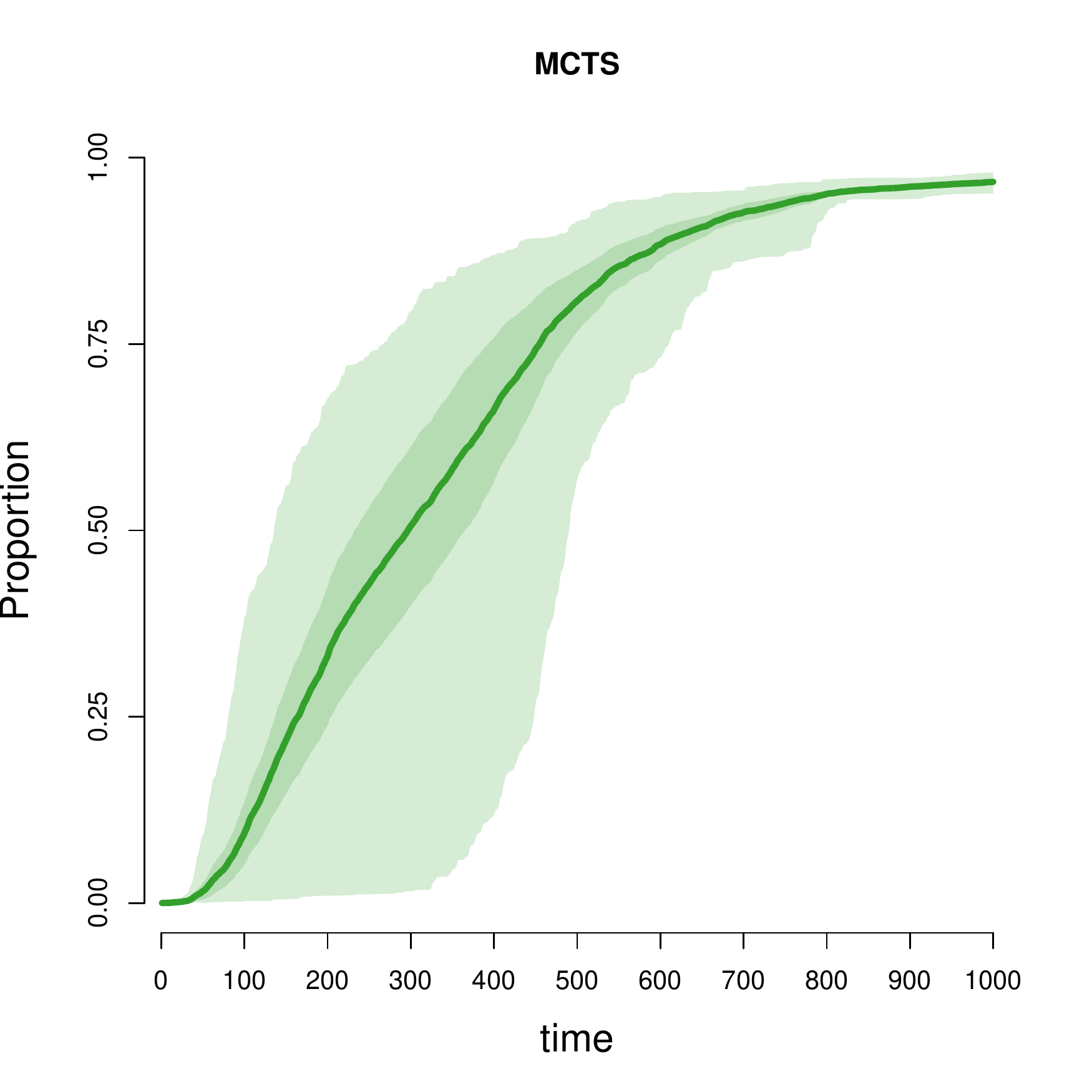}
\includegraphics[width=40mm]{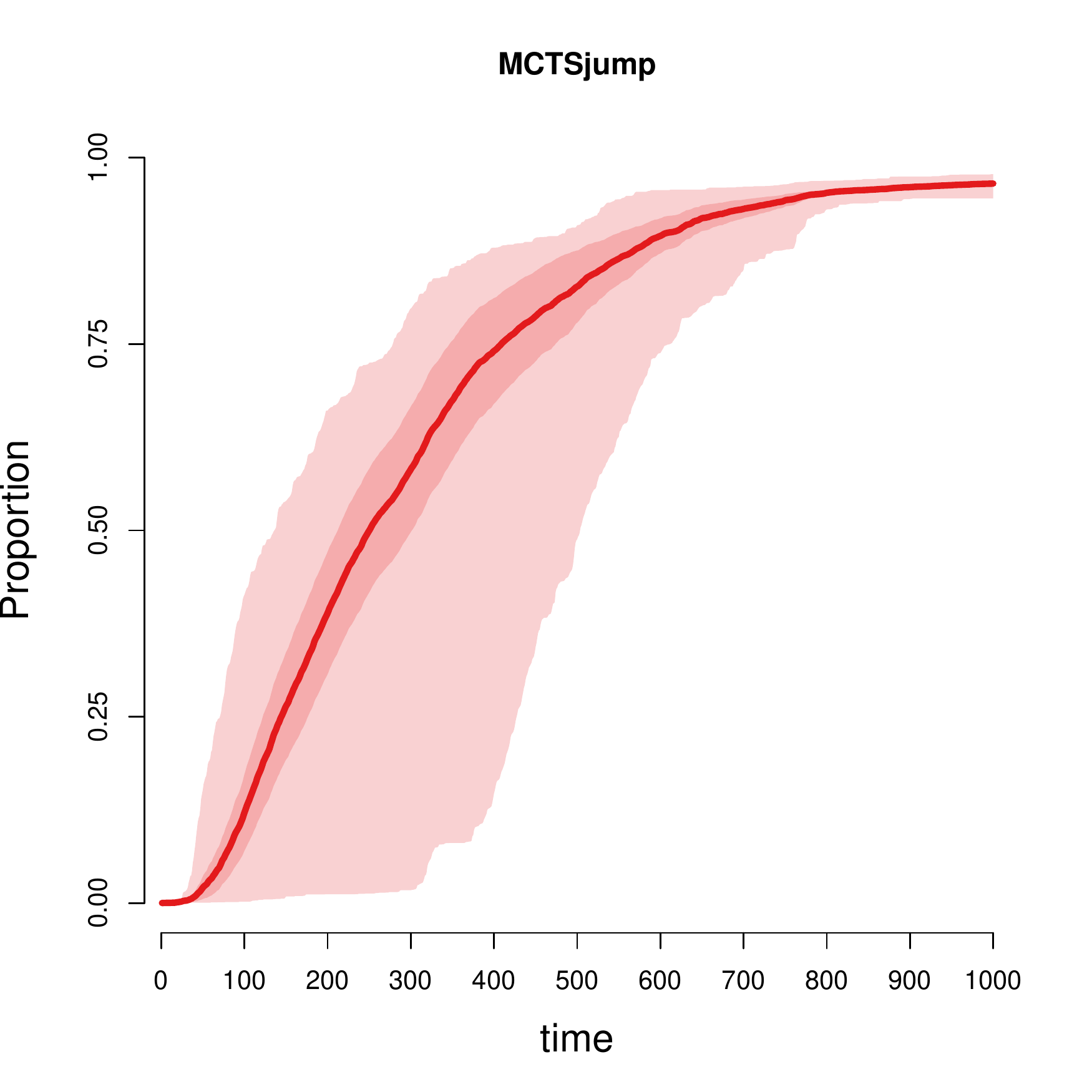} 
\includegraphics[width=40mm]{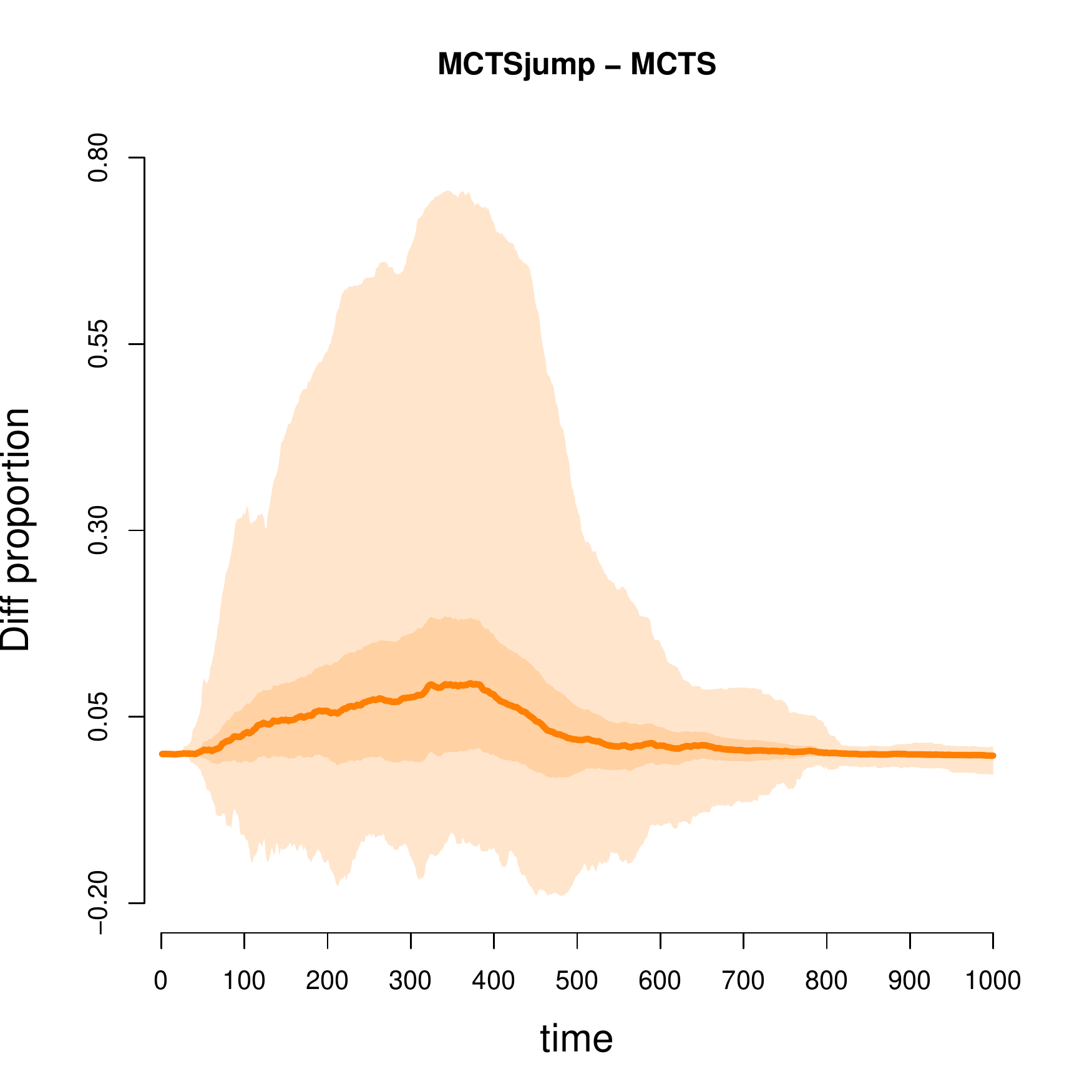}

\includegraphics[width=40mm]{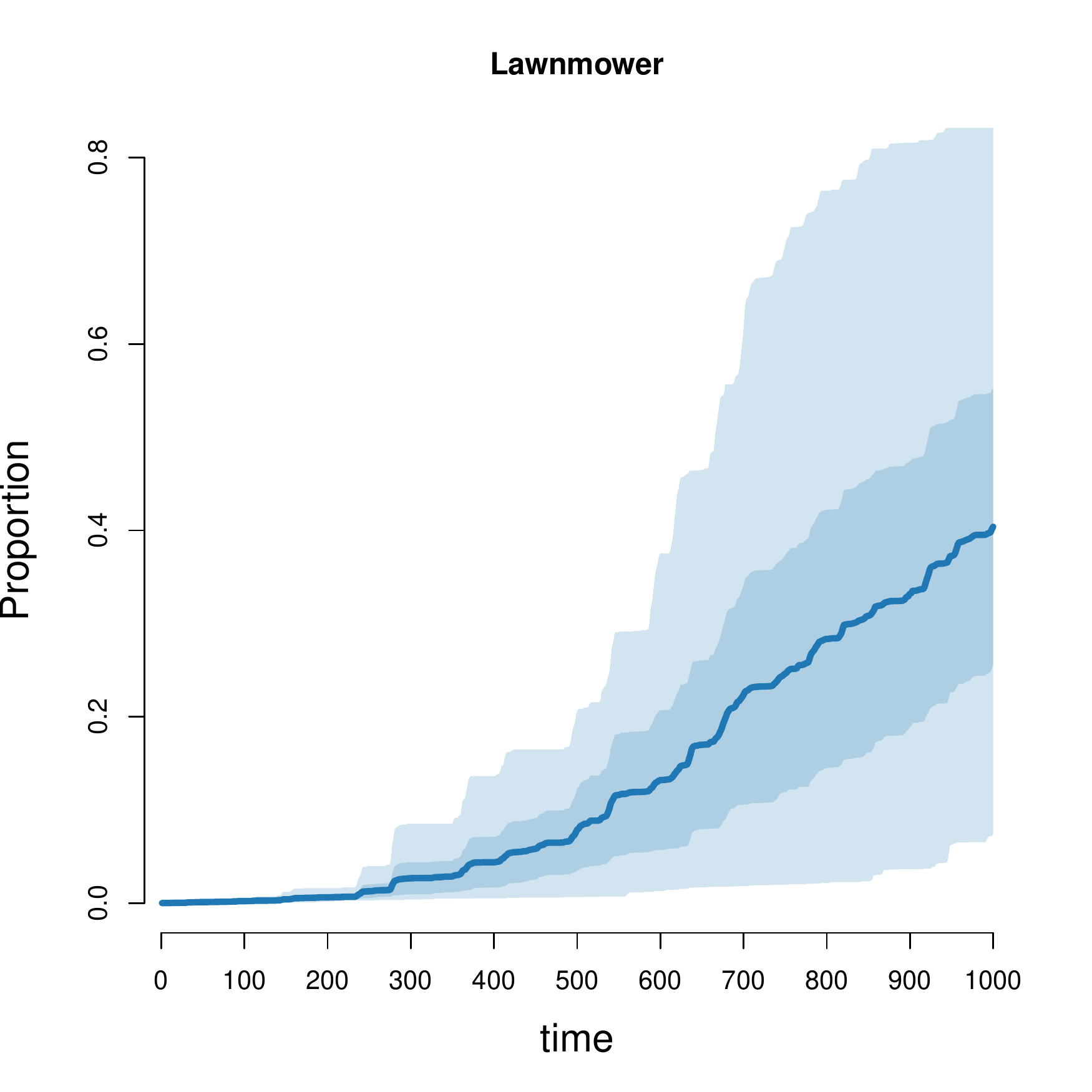}
\includegraphics[width=40mm]{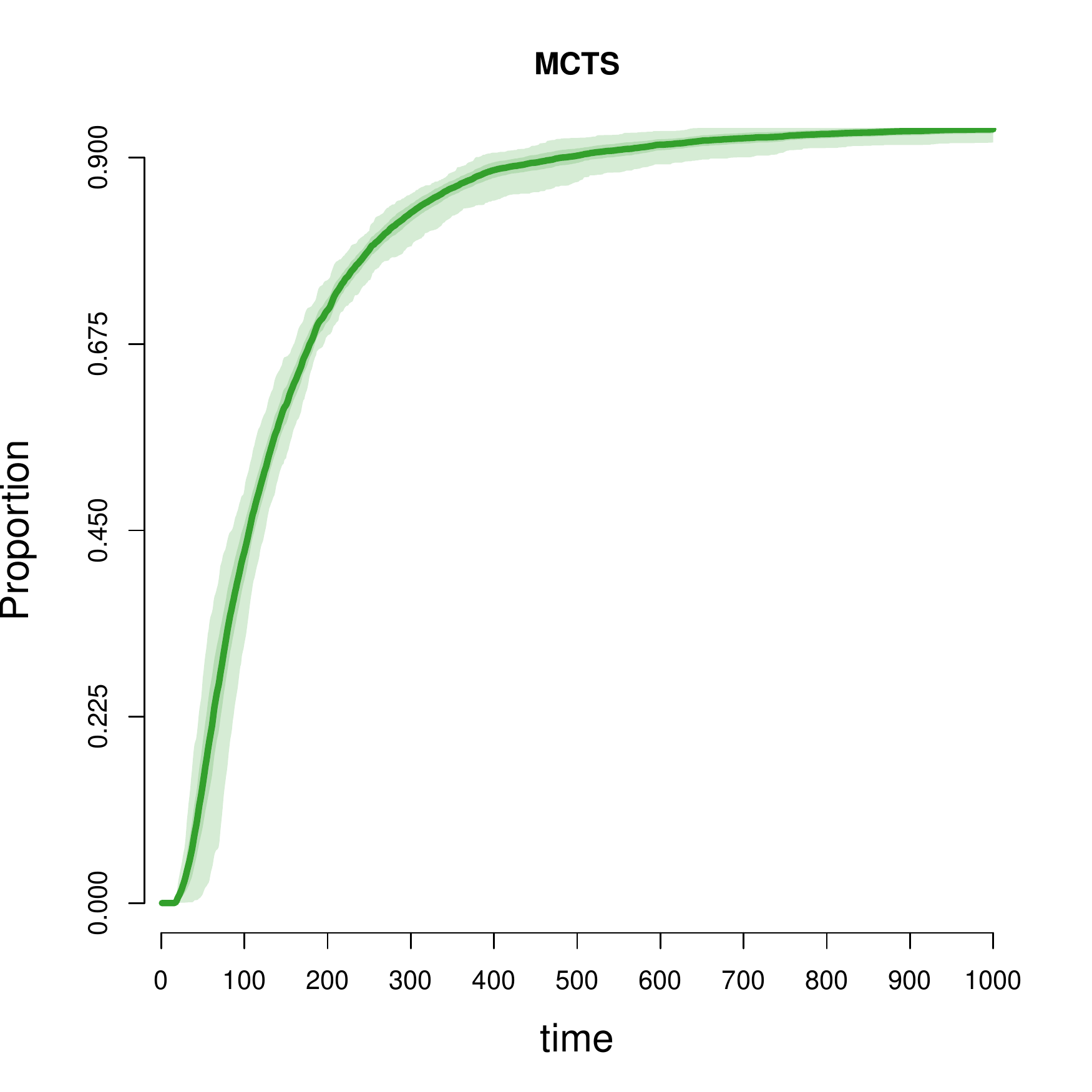}
\includegraphics[width=40mm]{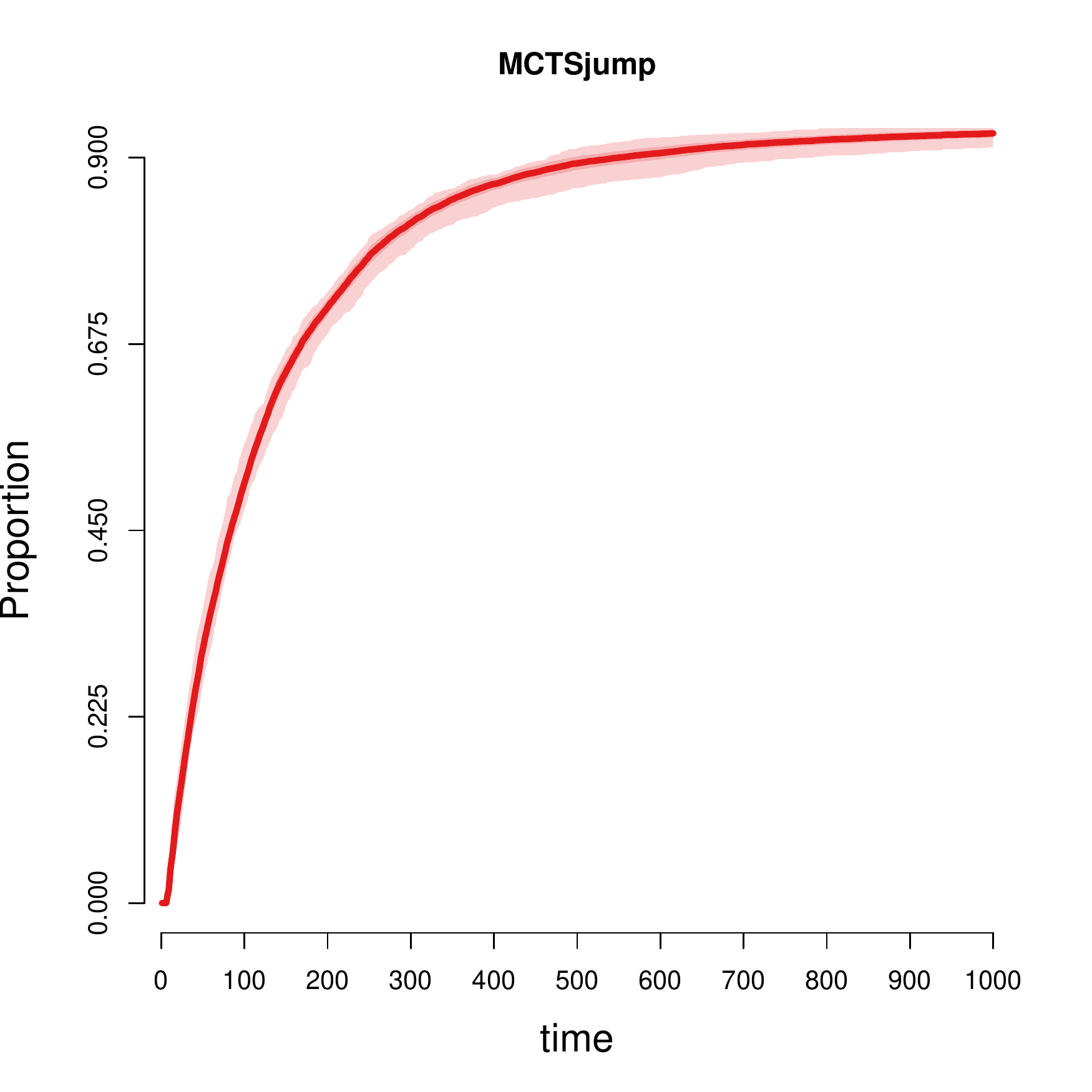}
\includegraphics[width=40mm]{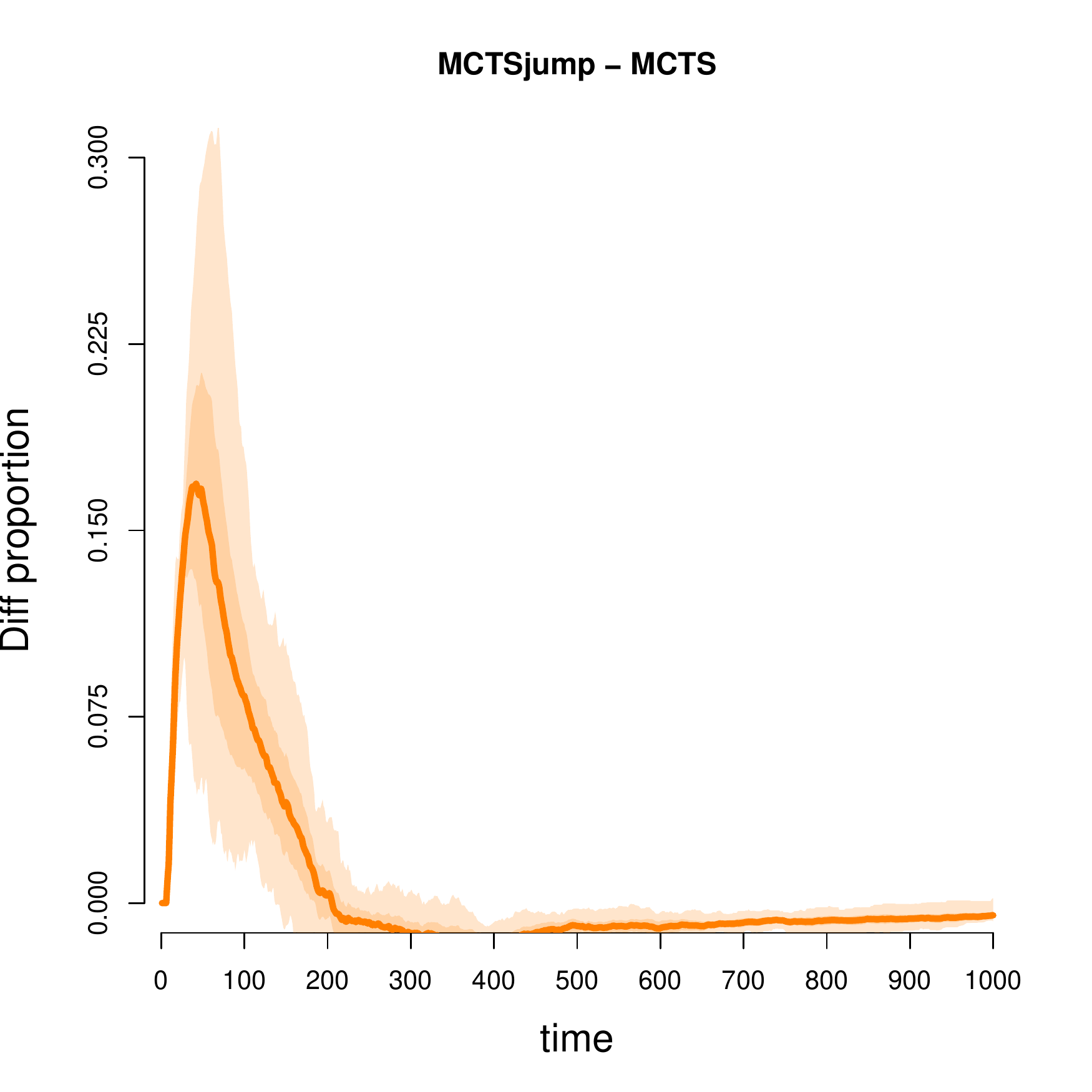}

\includegraphics[width=40mm]{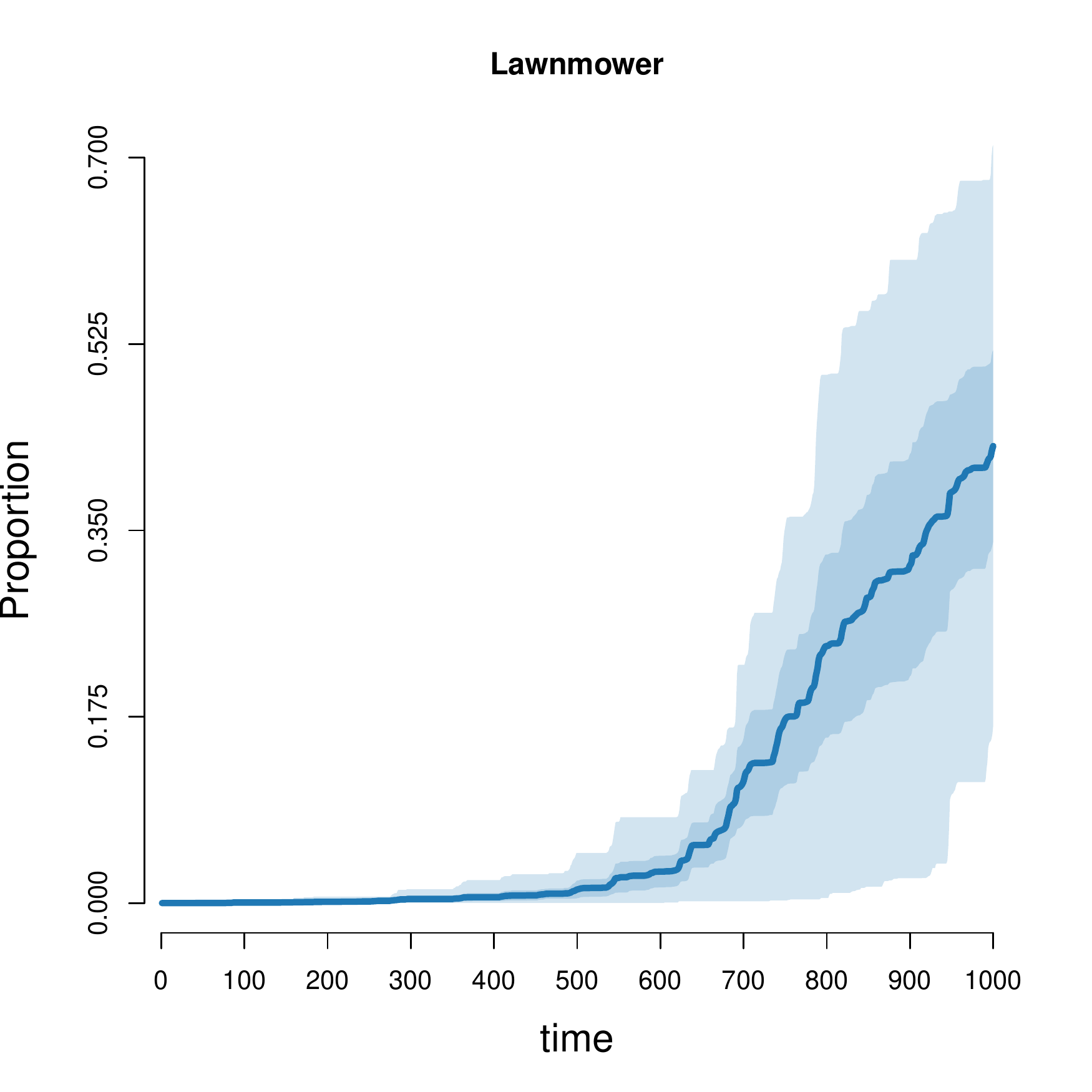}
\includegraphics[width=40mm]{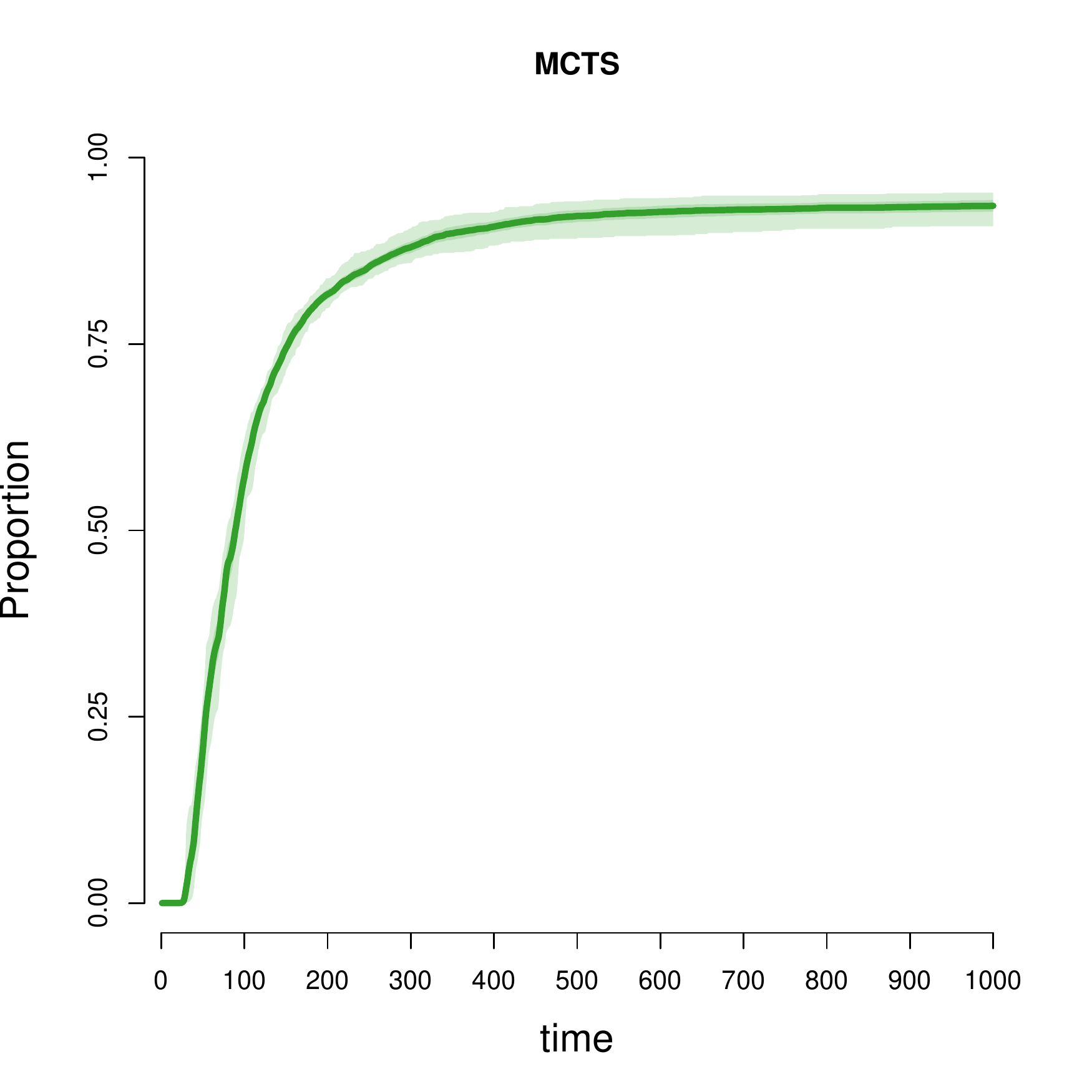}
\includegraphics[width=40mm]{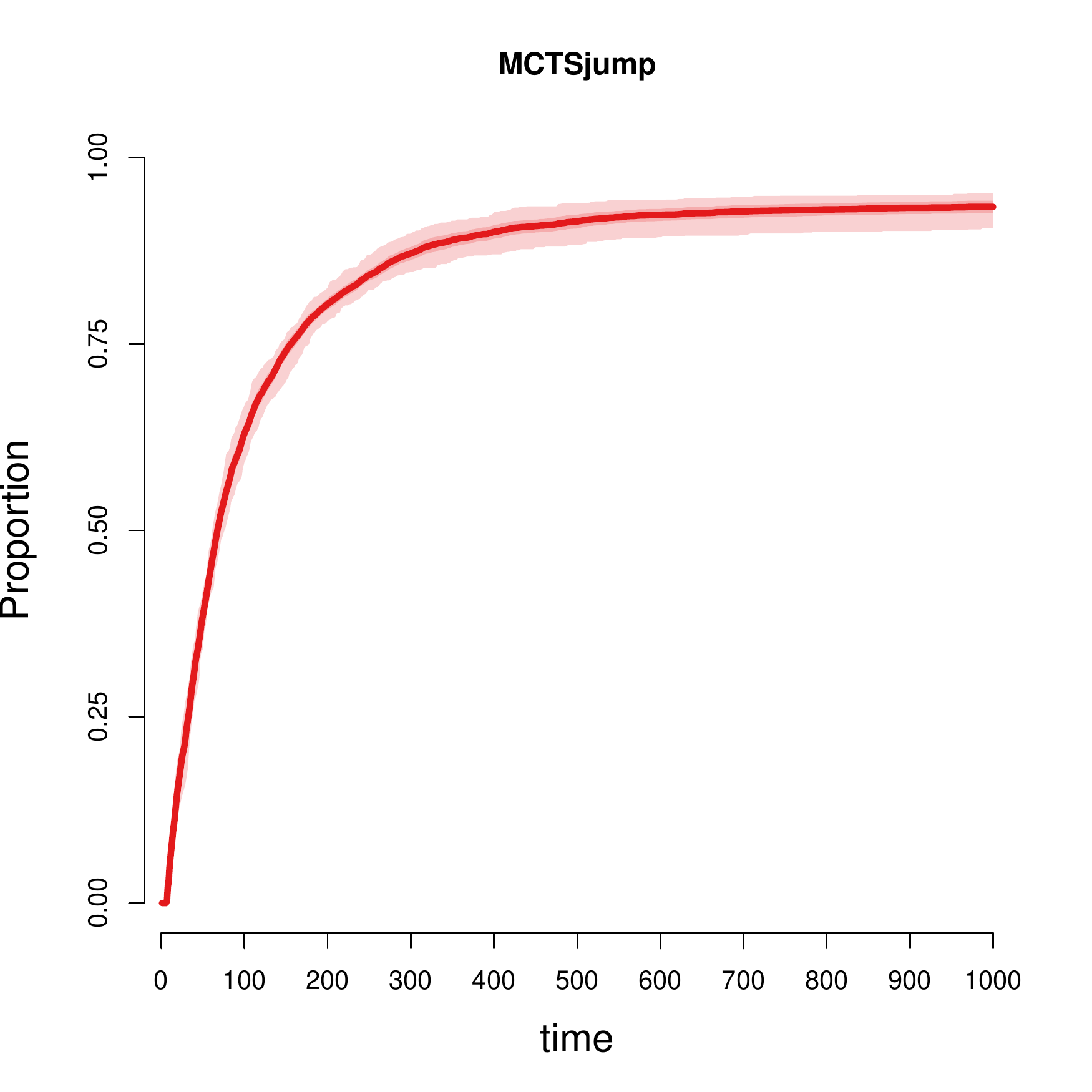}
\includegraphics[width=40mm]{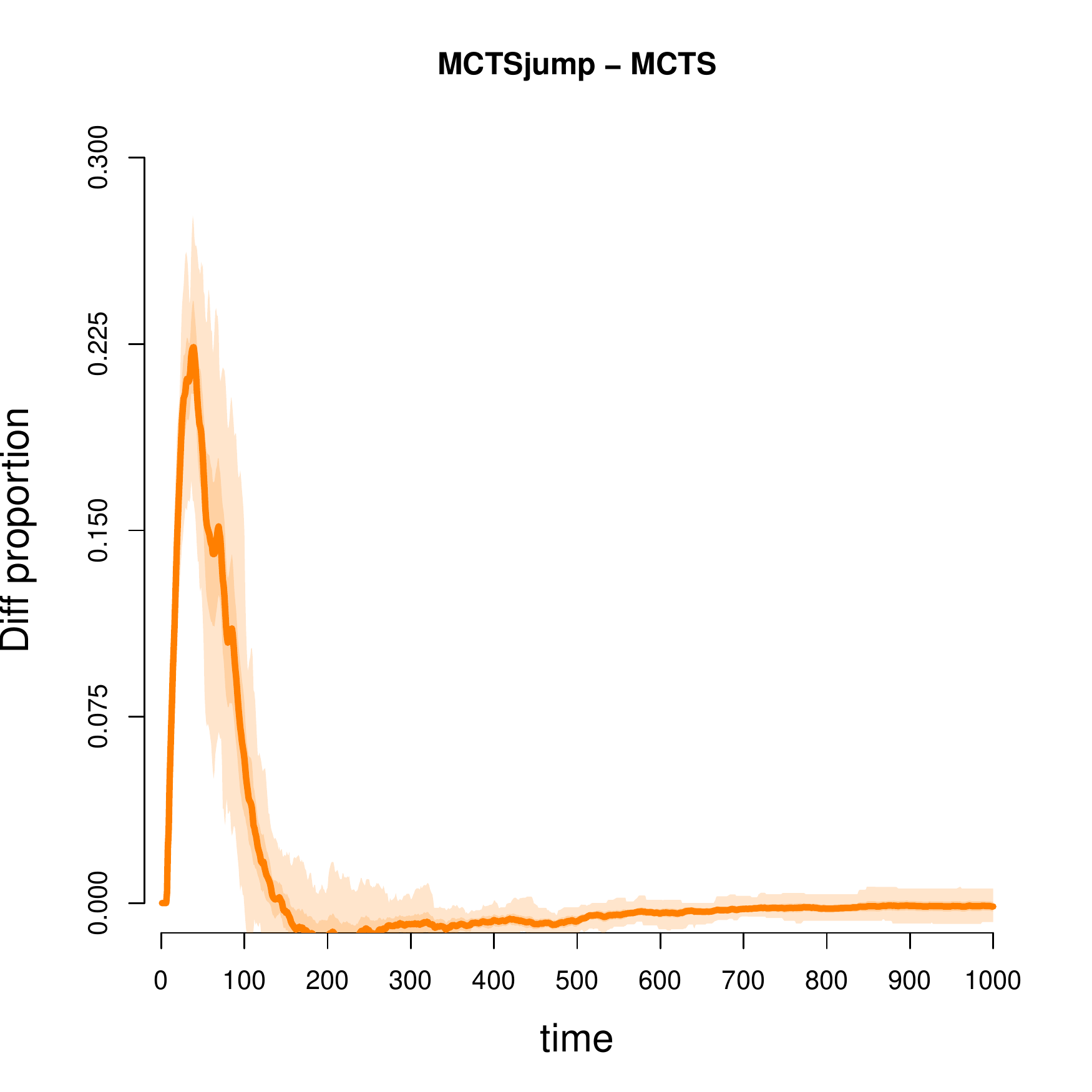}

\includegraphics[width=40mm]{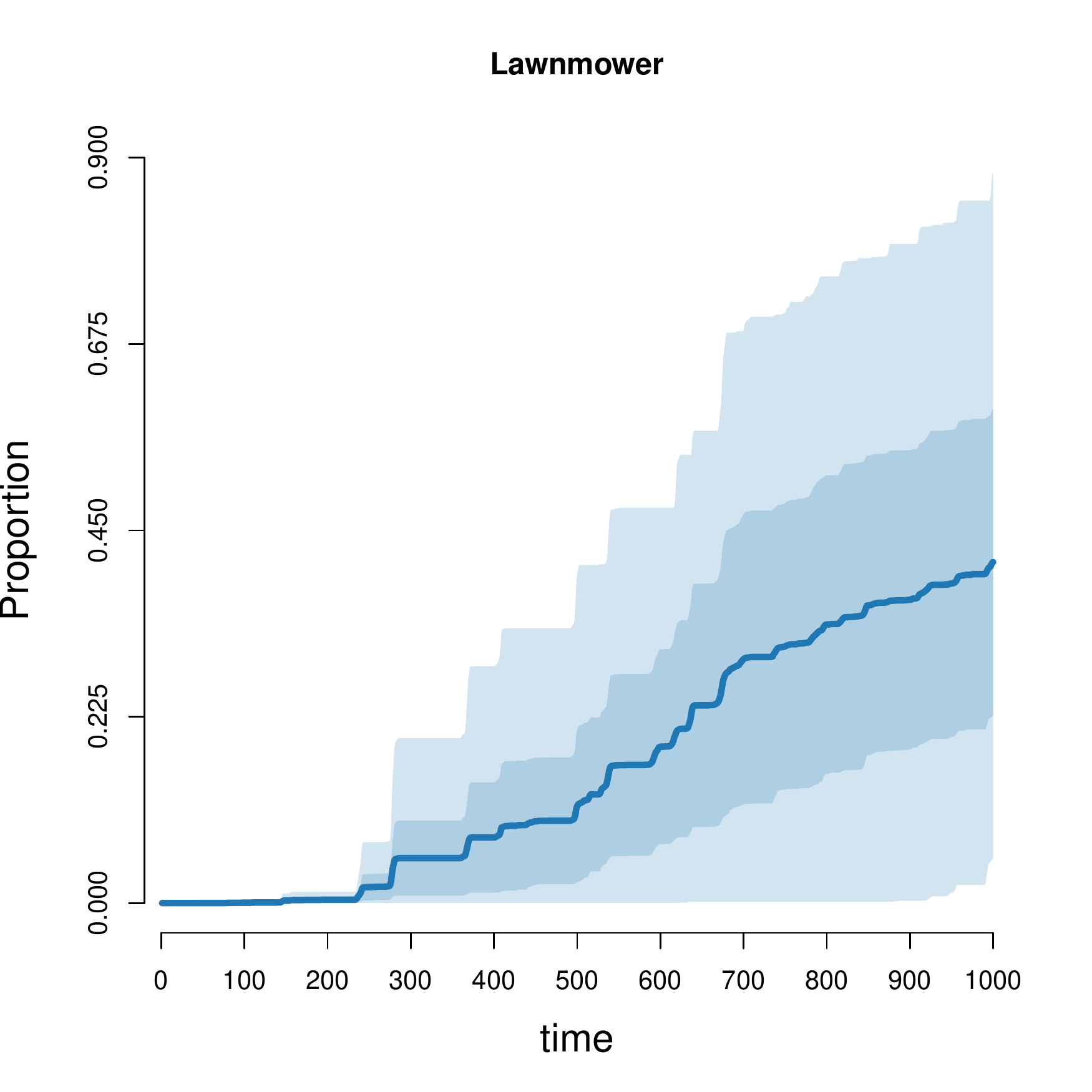}
\includegraphics[width=40mm]{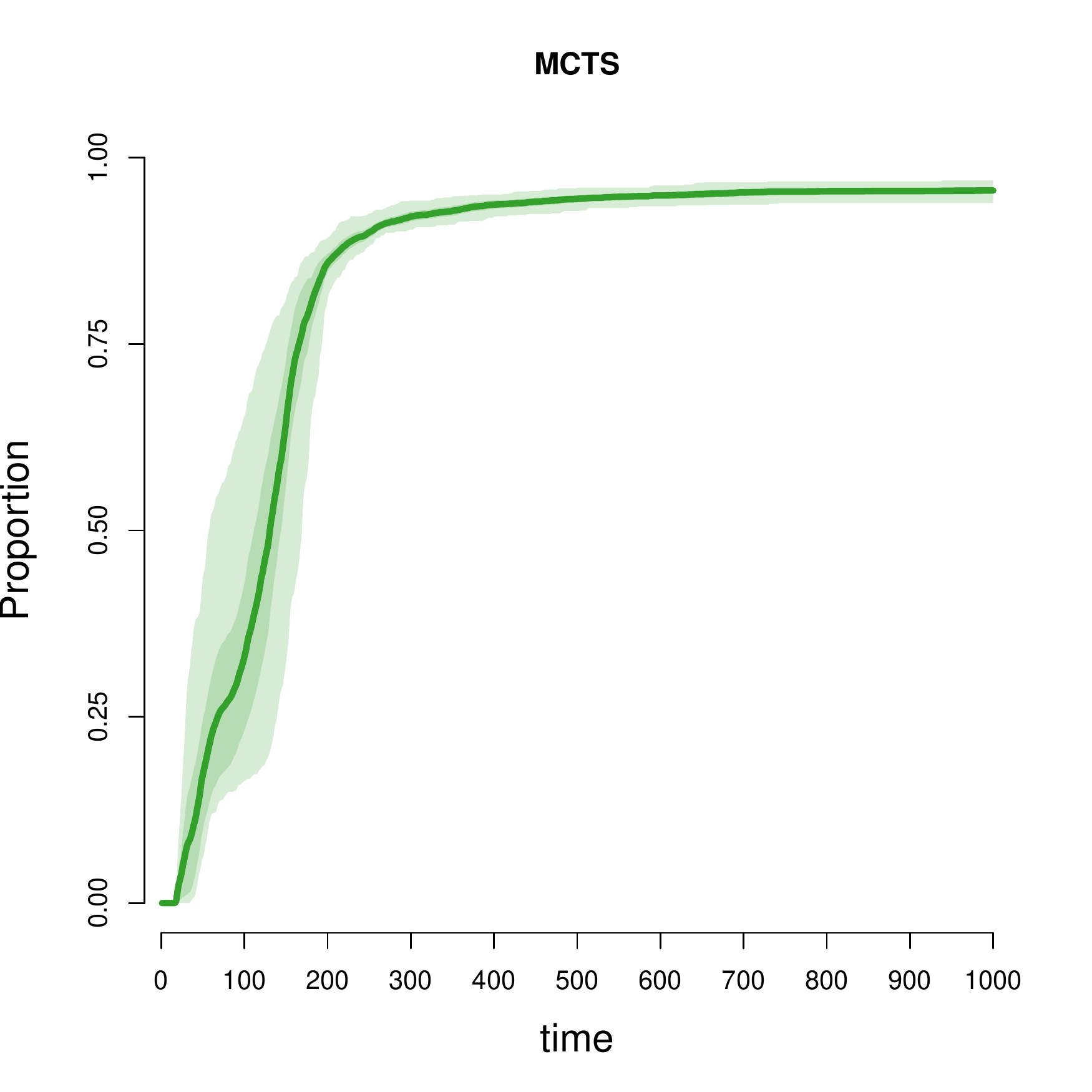}
\includegraphics[width=40mm]{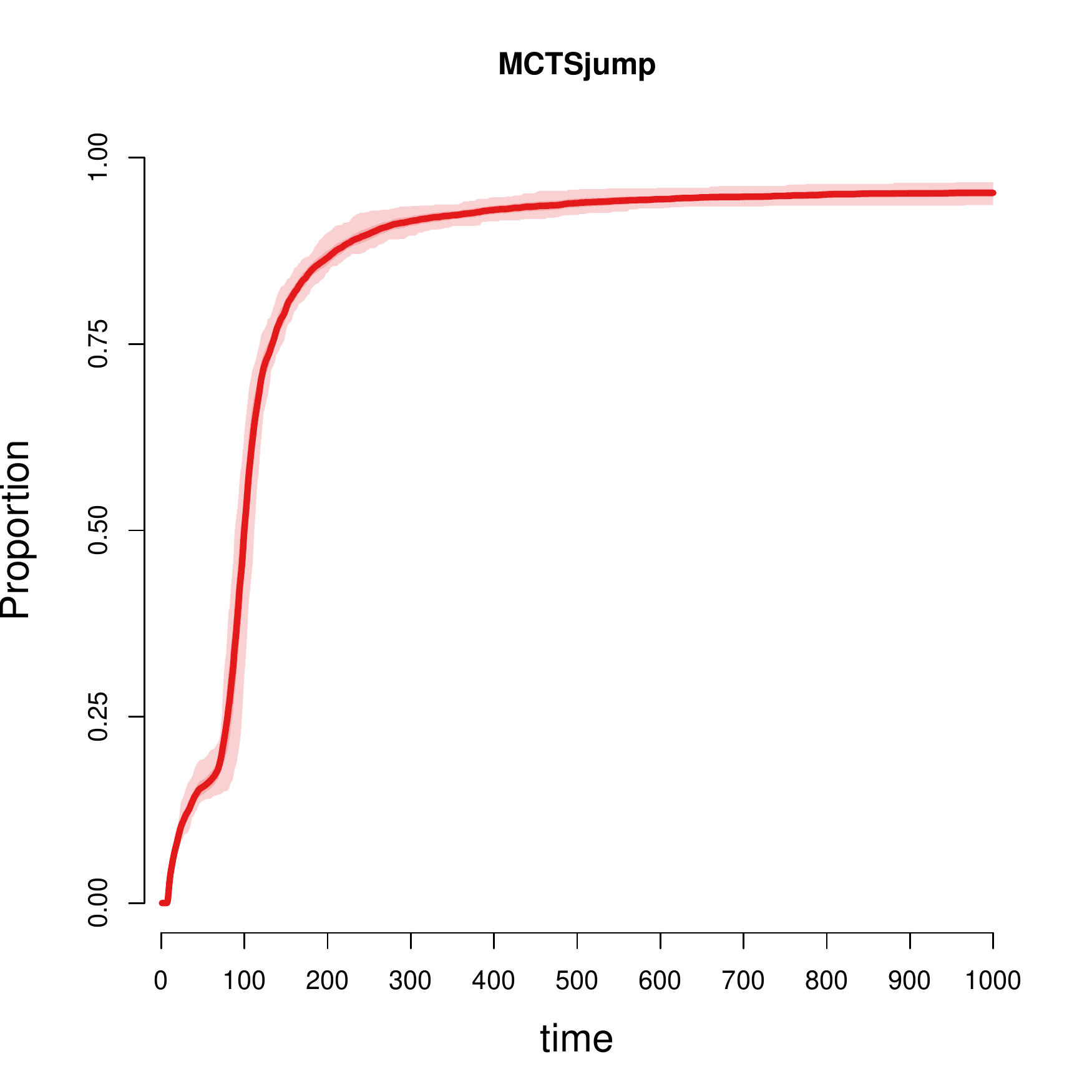}
\includegraphics[width=40mm]{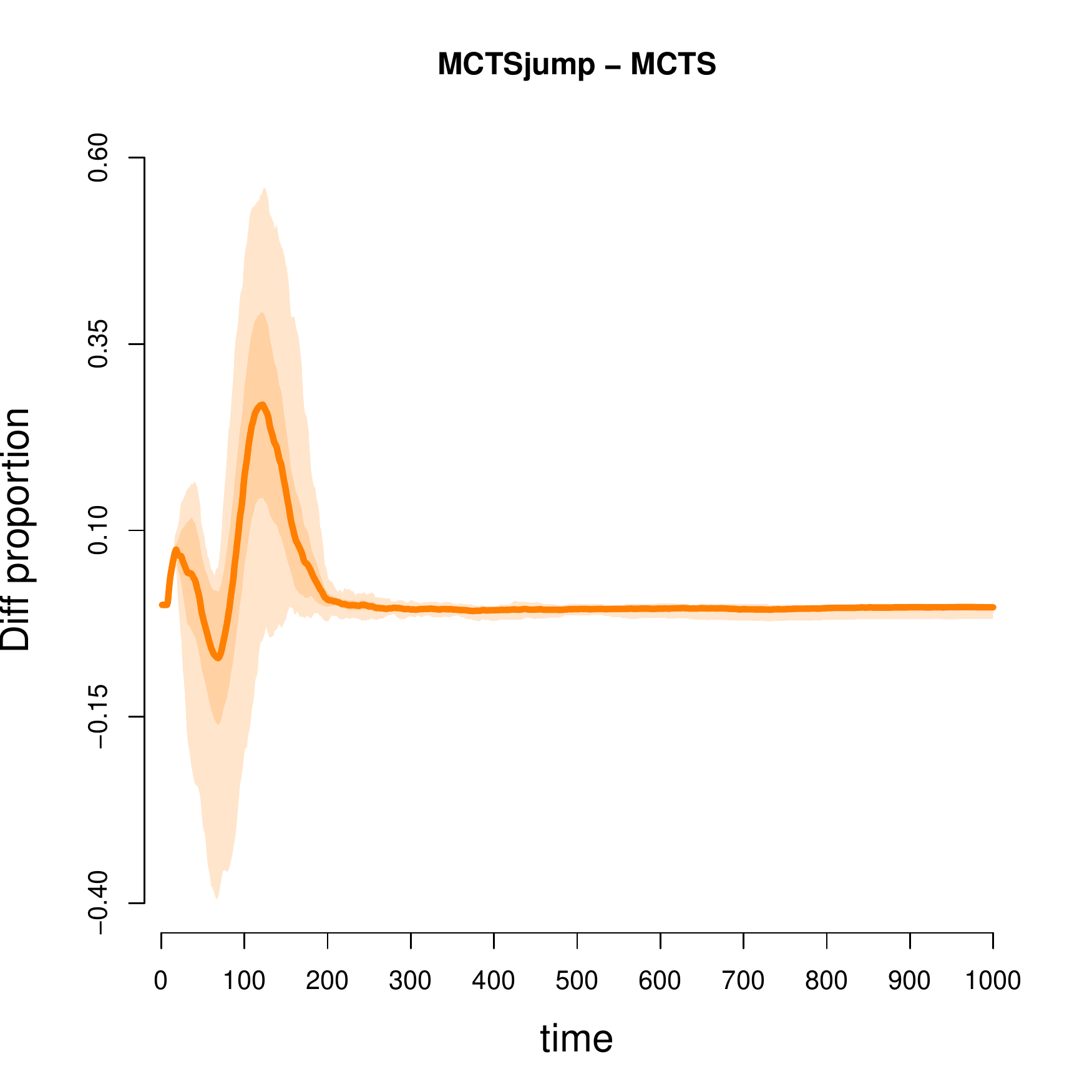}

\caption{Comparing the proportion of injured found as a function of search time (minutes) for the different strategies. The  rows correspond to each of the four scenarios A to D (top down). The graphs show the mean proportion of injured found as a function of search time (solid line) for the three strategies over 30 replicates, as well as the 95\% confidence bands for the mean (darker regions) and 95\% predictive bands for individual proportions in individual replicated datasets (lighter regions). The final column shows that same properties, but for the differences in proportions between MCTS and MCTSjump.}\label{fig:scenario2prop}
\end{figure*}

\vspace{-0.7em}
In this scenario there has been a localized terrorist attack, represented by a Gaussian in the injury field southwest of town. We show that using the proposed model, this can easily be encoded on the fly by first-responders, via e.g. a Gaussian spatial covariate prior centered on the reported site.  Figure~\ref{fig:scenario2prop} shows similar performance to Scenario B.

\subsection{Scenario D: Terrorist attack - one site unknown}
\begin{figure*}[h!] 
\includegraphics[trim=25mm 8mm 25mm 8mm,width=55mm]{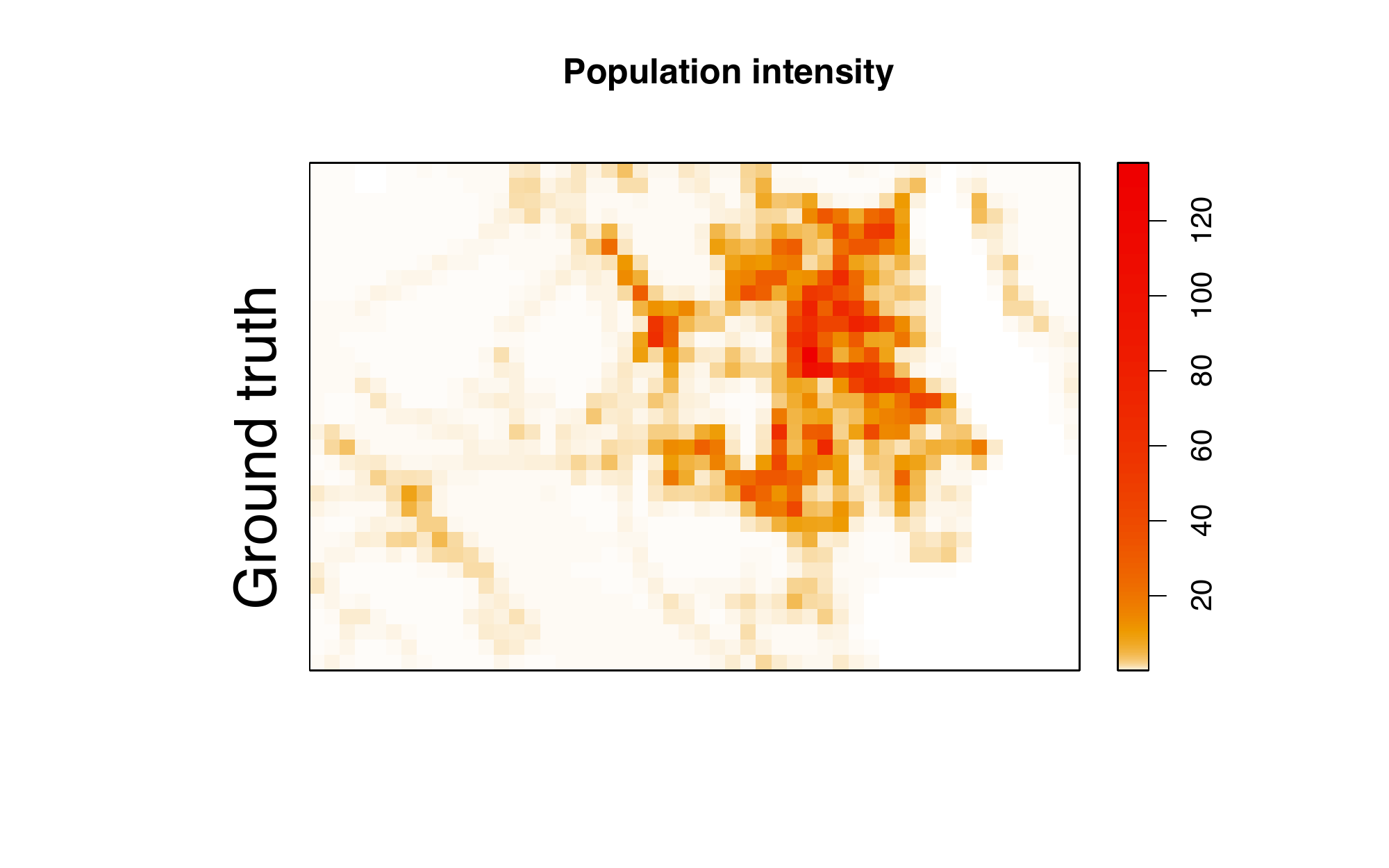}
\includegraphics[trim=25mm 8mm 25mm 8mm,width=55mm,clip]{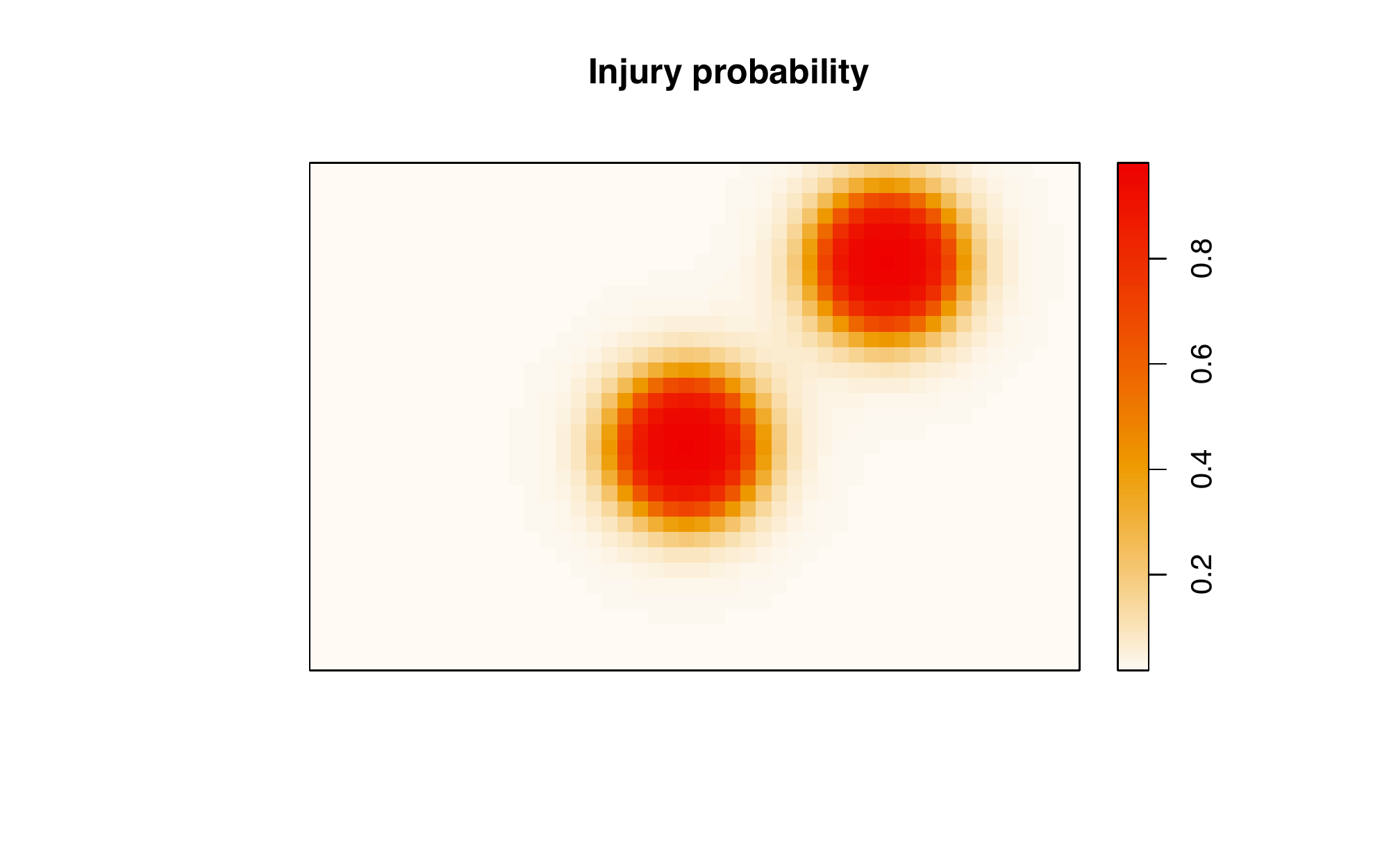}
\includegraphics[trim=25mm 8mm 25mm 8mm,width=55mm,clip]{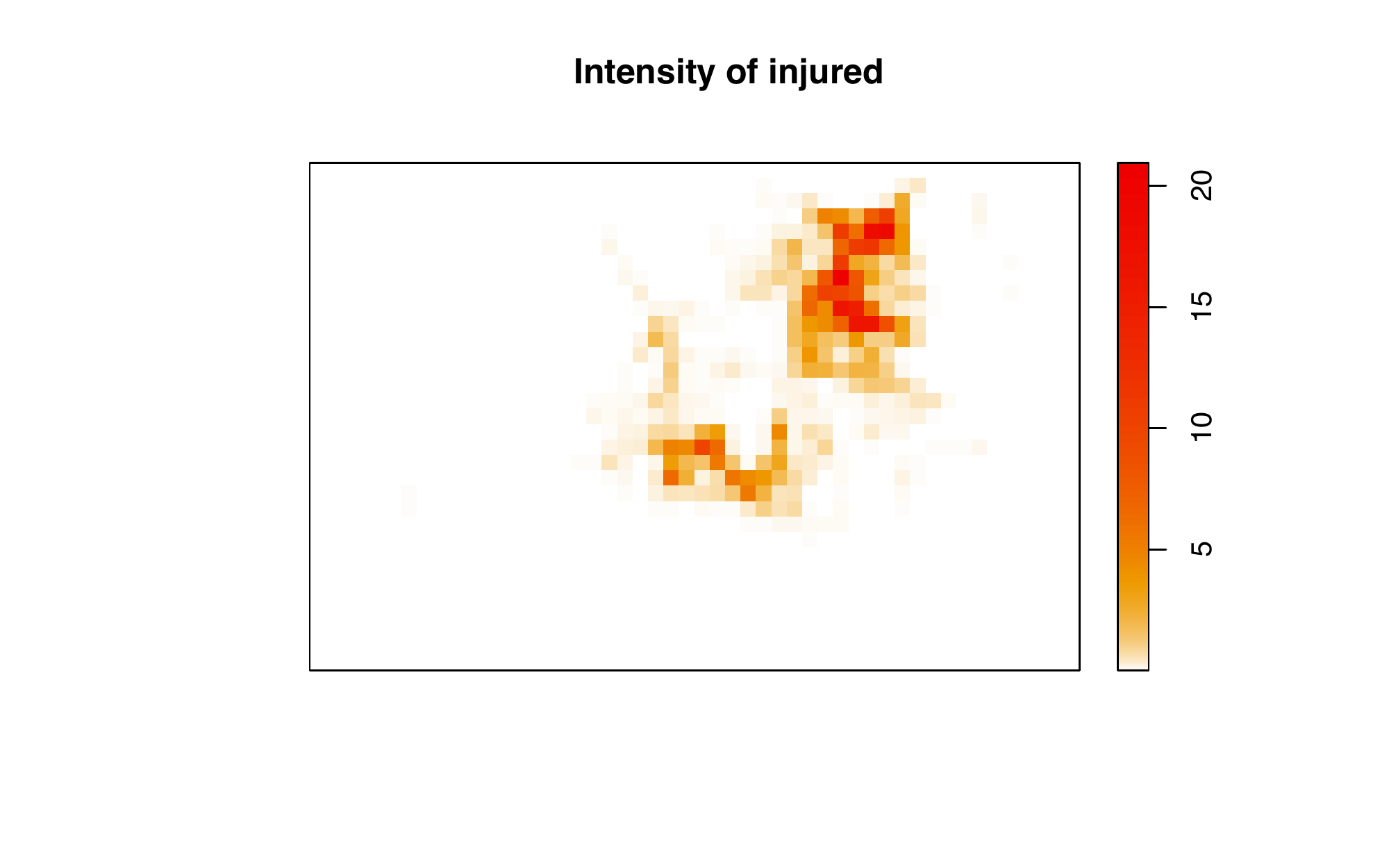}\newline
\includegraphics[trim=25mm 8mm 25mm 5mm,width=55mm]{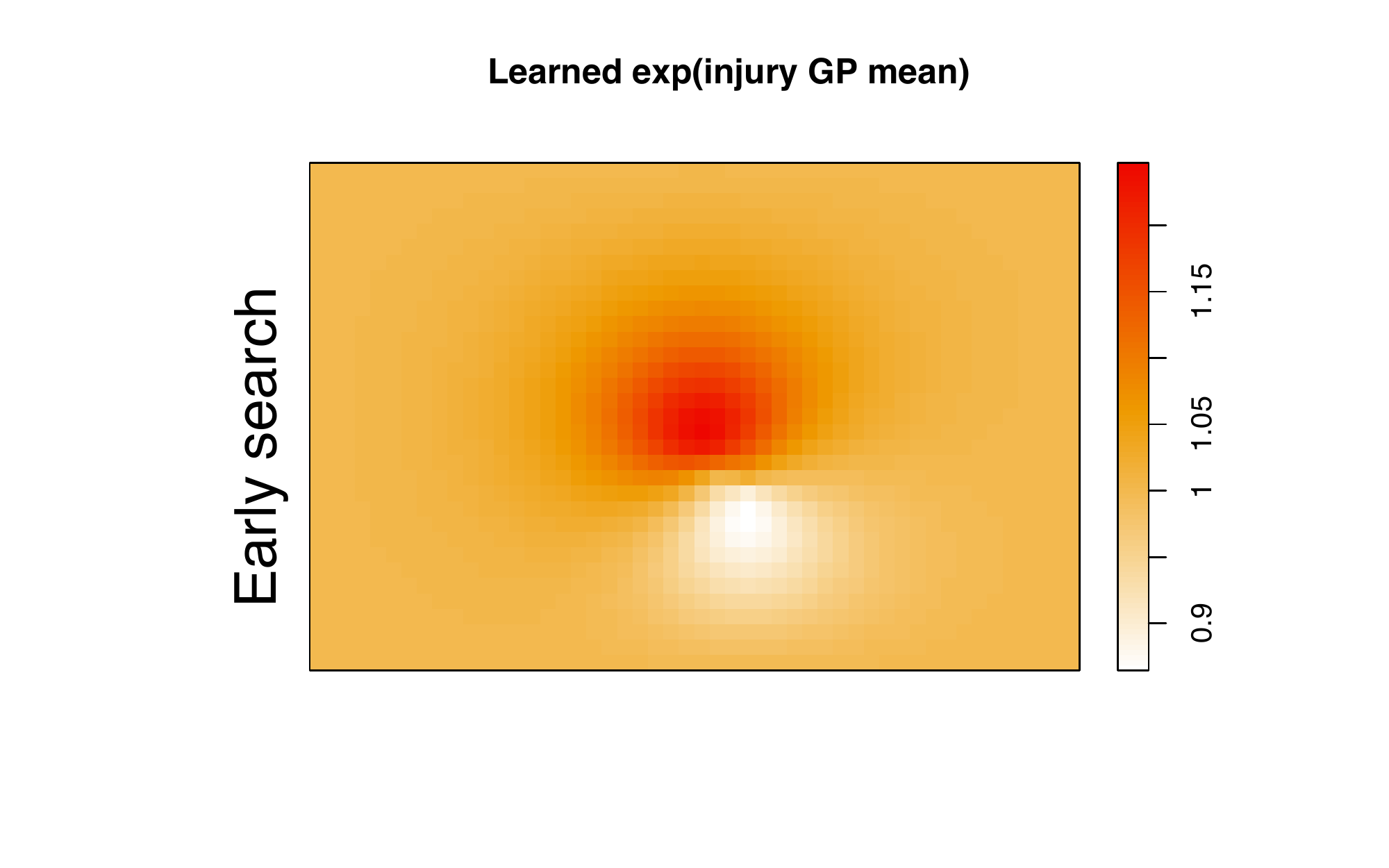}
\includegraphics[trim=25mm 8mm 25mm 5mm,width=55mm,clip]{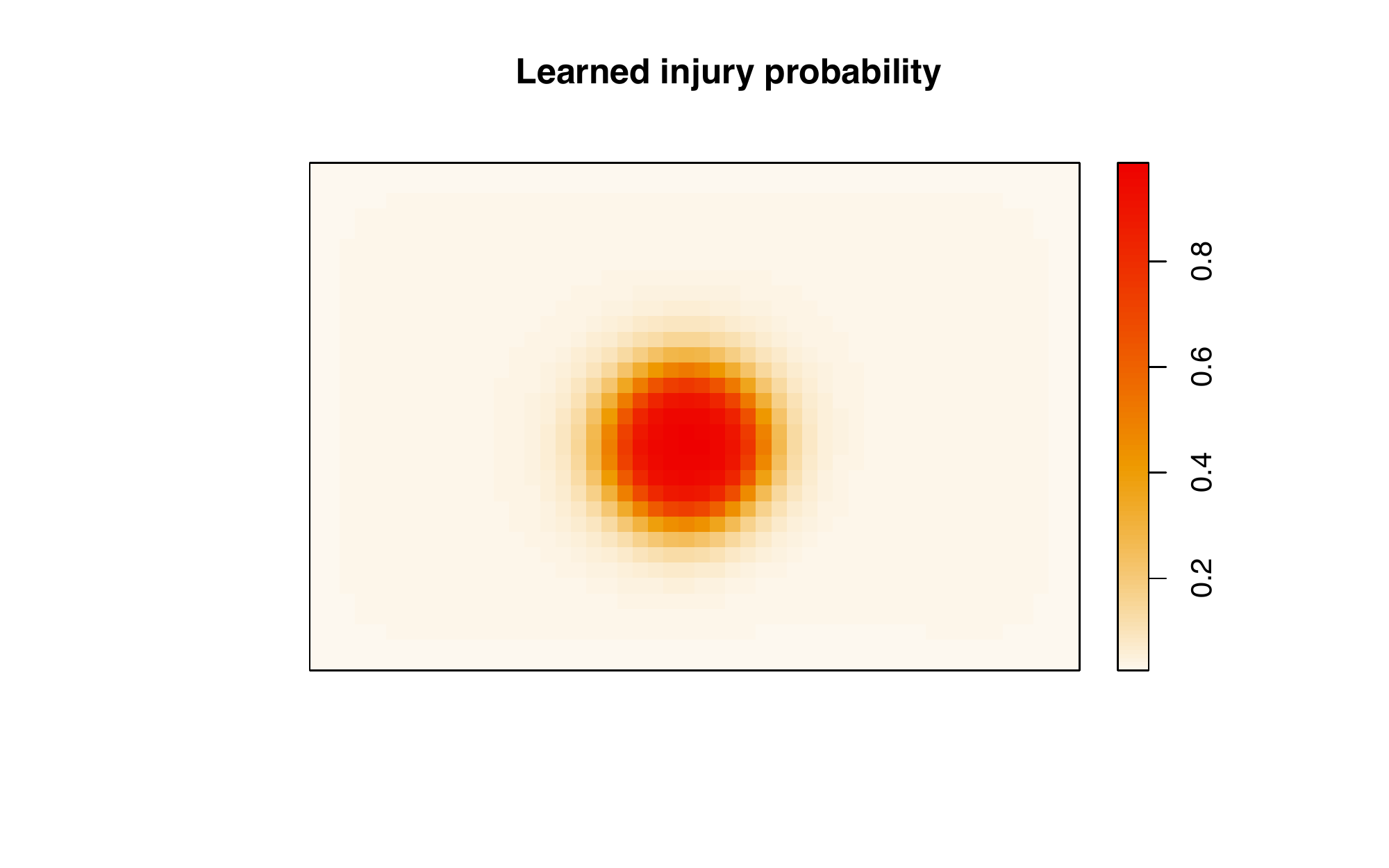}
\includegraphics[trim=25mm 8mm 25mm 5mm,width=55mm,clip]{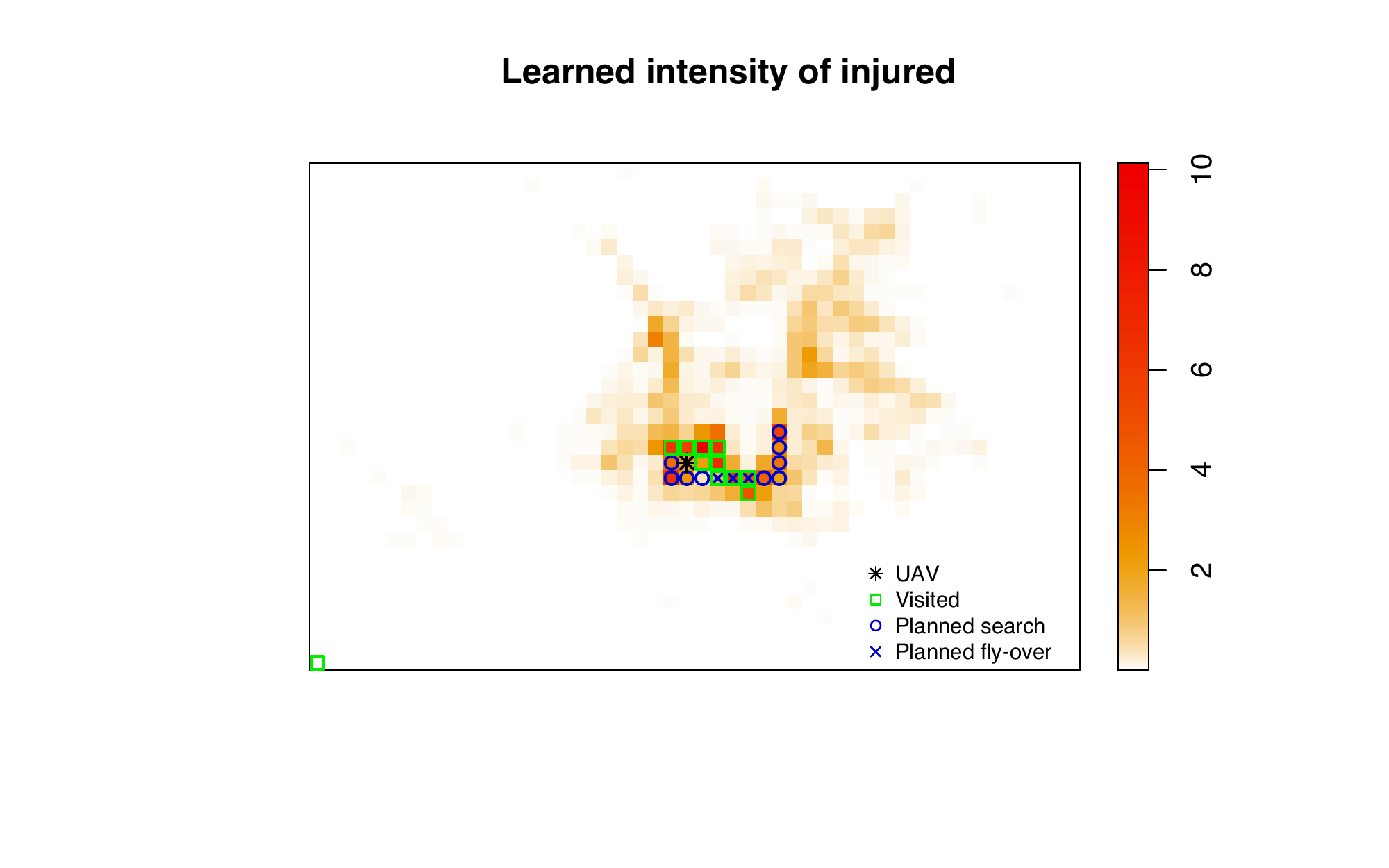}\newline
\includegraphics[trim=25mm 8mm 25mm 5mm,width=55mm]{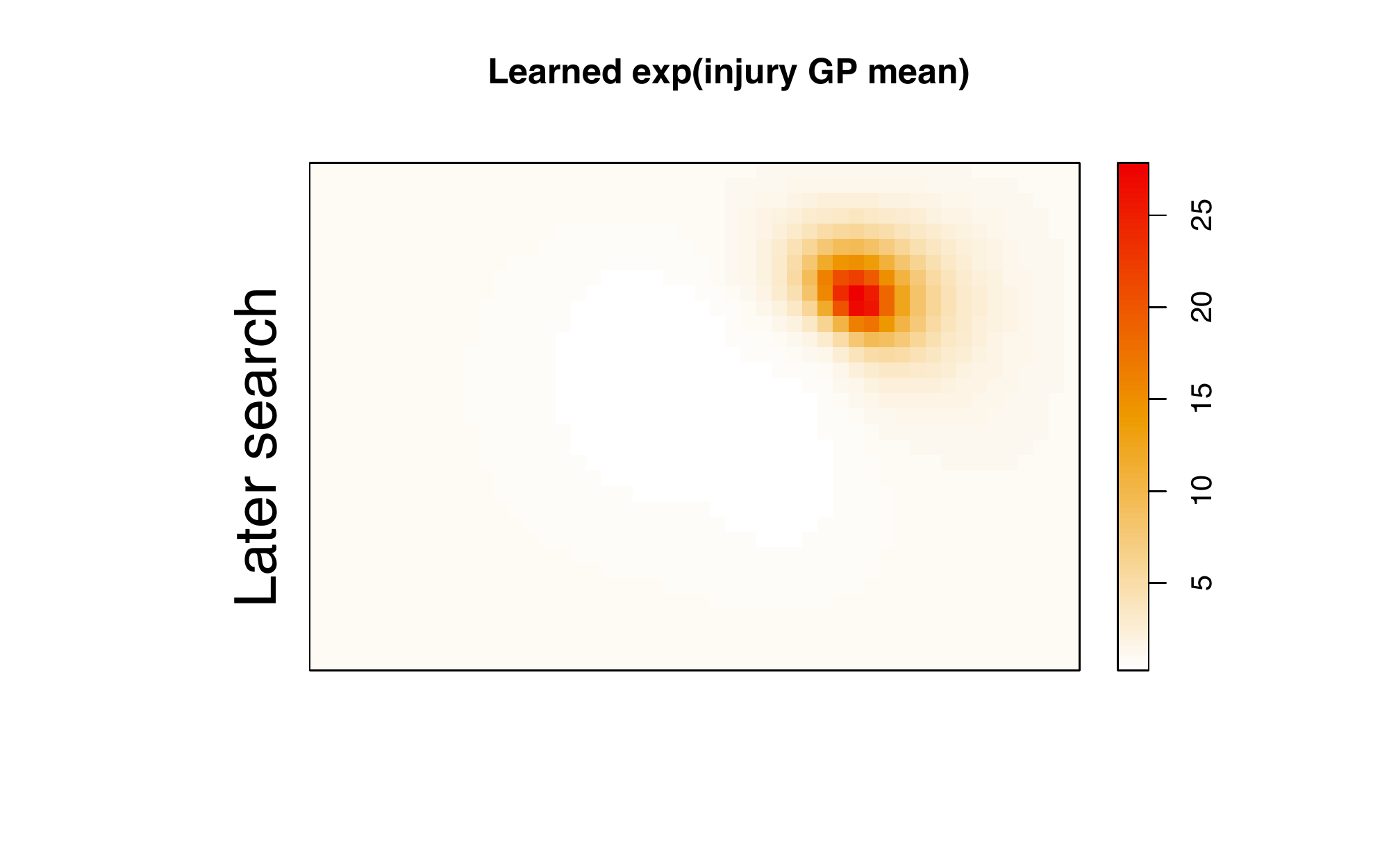}
\includegraphics[trim=25mm 8mm 25mm 5mm,width=55mm,clip]{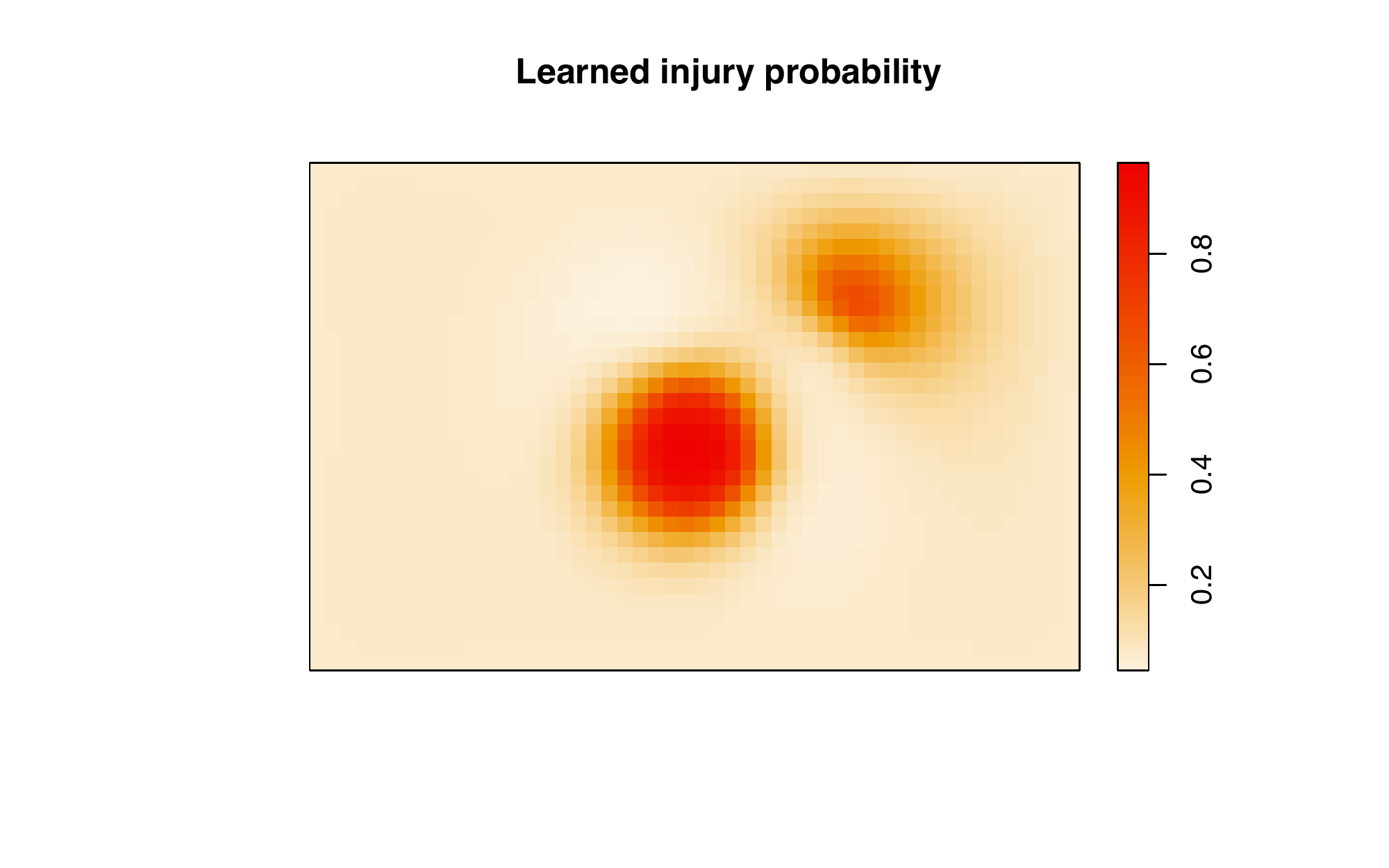}
\includegraphics[trim=25mm 8mm 25mm 5mm,width=55mm,clip]{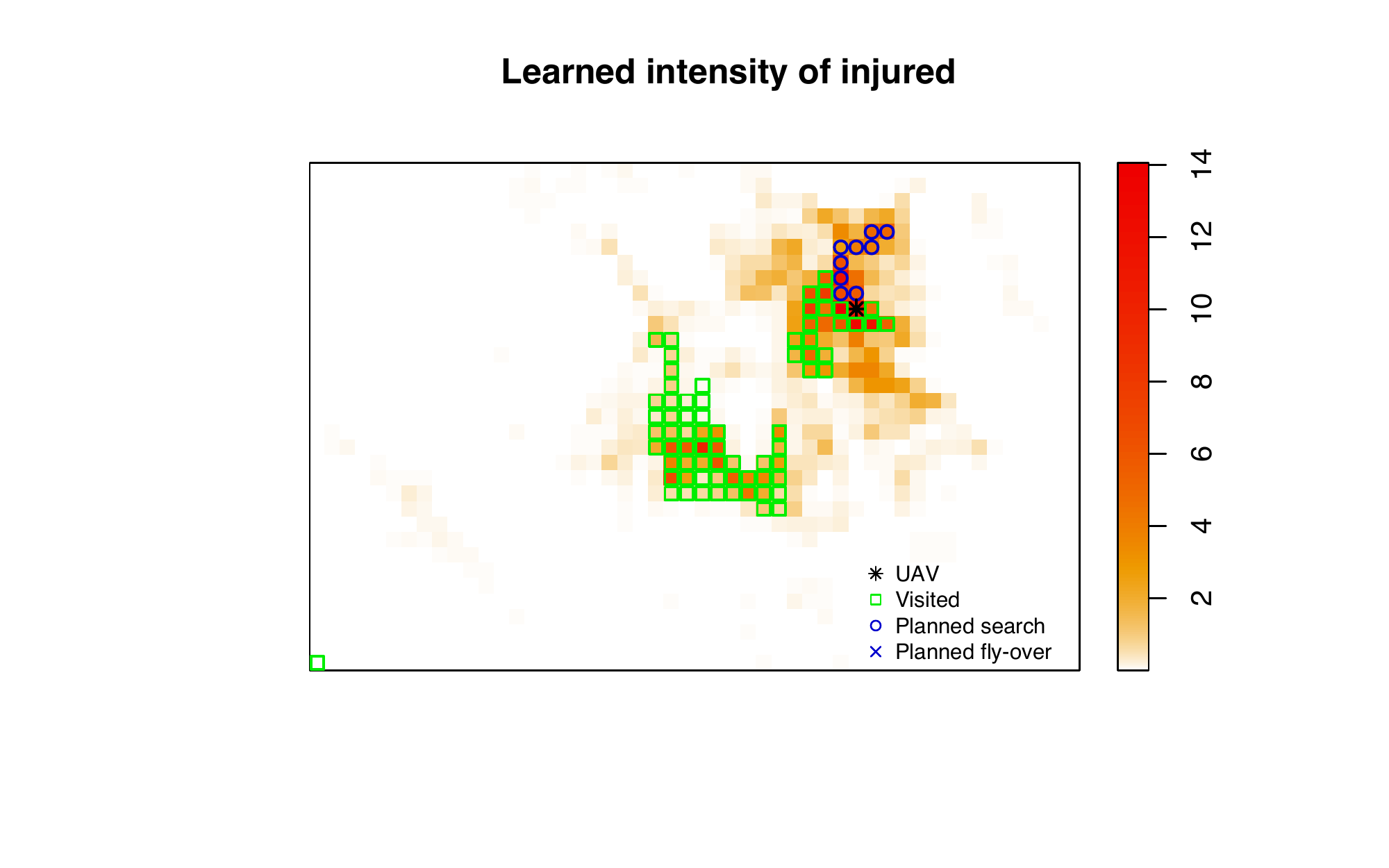}
\caption{True and inferred population and injury maps for scenario 6. All intensity plots are based on detectable persons only. The first row displays the ground truth used for simulating the data, showing the expected number of persons per cell, the probability of being injured in each cell, and the expected number of injured in each cell. The second row shows $\exp(E(\xi_q))$,$E(q)$, and  $E(\lambda r q)$, with expectations with respect to the posterior after 14 search iterations. The third row is equivalent to the second row, but after 91 iterations of search.}\label{fig:scenario6maps}
\end{figure*}

\vspace{-0.7em}
Finally, we showcase all the capabilities of the structured model by extending Scenario C. In this case, there is a terrorist attack with one site encoded by a Gaussian spatial prior. However, early information during catastrophes is often incomplete. In this case there is also a second attack site unknown to us. The results show the model quickly picks up on this. In particular, \texttt{MCTSJump} flies to and explores the a priori known site, then without further information will jump around and explore high population areas. At some point it stumbles on injured near the second site, the spatial field quickly learns the local anomaly in injury probability, and the planner focuses on that area. A simulation run of Scenario D is shown in the supplementary video material\footnote{\url{https://youtu.be/wyD0O5hF5tE}} and Figure~\ref{fig:scenario6maps}.\looseness=-1000

\section{Conclusions}
We present a new framework for search-and-rescue based on real-time learning and decision making with a hierarchically structured spatial point process. The model is built from spatially referenced components on which there is usually ample prior information in search-and-rescue problems: i) the distribution of persons, ii) the probability of detecting a person, and iii) the probability of injury. Learning spatial processes and acting on them in real-time is a hard problem. We propose a novel combination of approximate Bayesian learning using INLA combined with a MCTS strategy adapted to the search problem.

We assess the empirical performance of the method on several simulated scenarios on a real map with publicly available GIS data, and show that prior information can be very efficiently used in our model to clearly outperform a conventional search strategy. We also demonstrate that the spatial fields can fill in for missing prior information in a very adaptable manner.

The framework proposed here can be extended in many interesting directions, for example to dynamic problems where the intrinsic state variables evolve over time, such as disasters involving gas leakage or a rescue operation at sea. It would be interesting to generalize the model and the approximate inference method to other data distributions and other link functions than the exponential and logistic. While this work indicates that using just one UAV very cleverly can make a large difference, in future work we also intend to extend it to search with a team of real UAV.
\vfill


\bibliographystyle{apalike}
\bibliography{ref}

\vfill
\newpage
\onecolumn
\section*{\centering\huge Supplementary Material}
\vfill
\vfill

\begin{figure*}[h!]
\vspace{1in}
\includegraphics[width=40mm]{figs/injuredFoundScenario3lawnmower.pdf}
\includegraphics[width=40mm]{figs/injuredFoundScenario3mcts.pdf}
\includegraphics[width=40mm]{figs/injuredFoundScenario3mcts_jump.pdf} 
\includegraphics[width=40mm]{figs/diffInjuredFoundScenario3mcts_jump_vs_mcts.pdf}
\includegraphics[width=40mm]{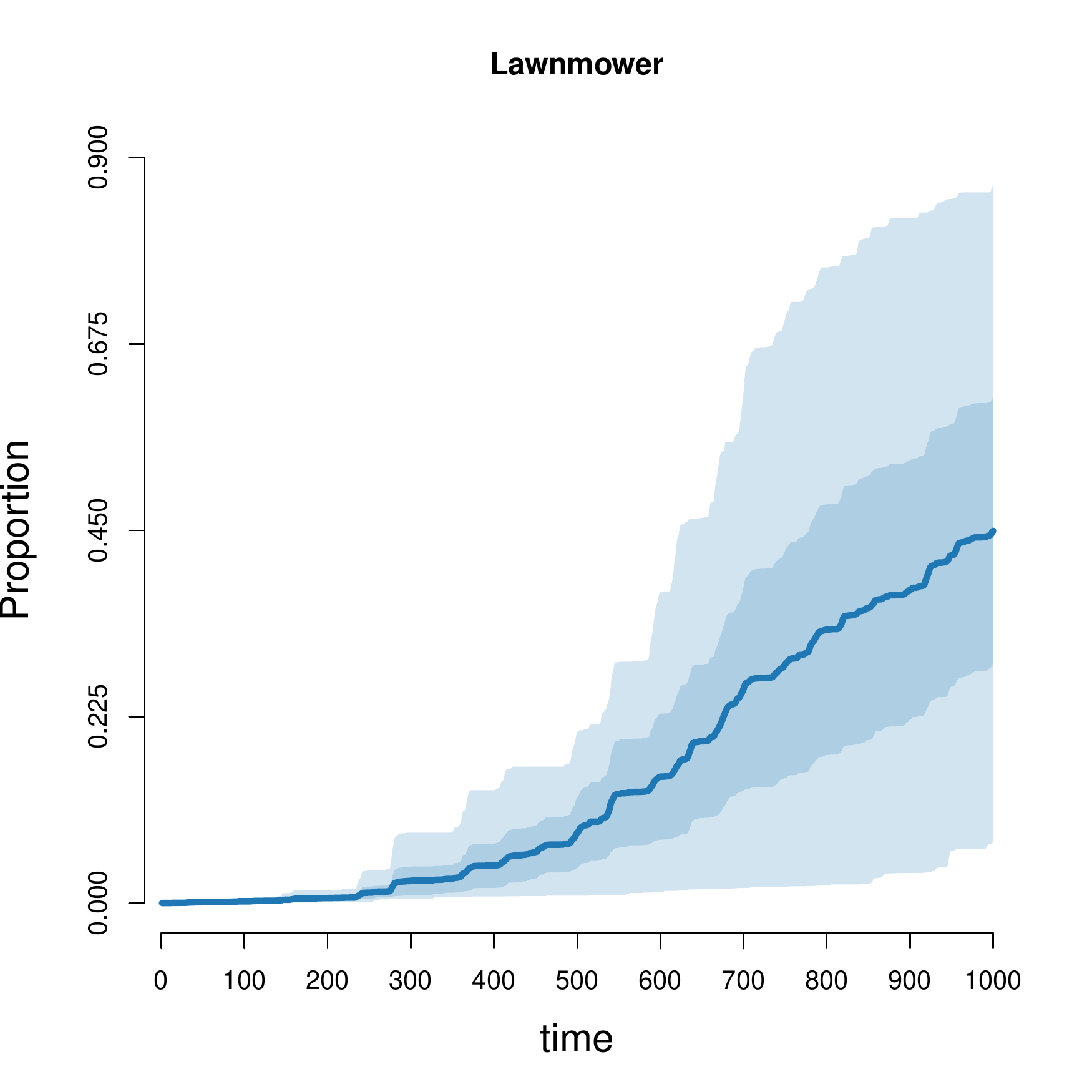}
\includegraphics[width=40mm]{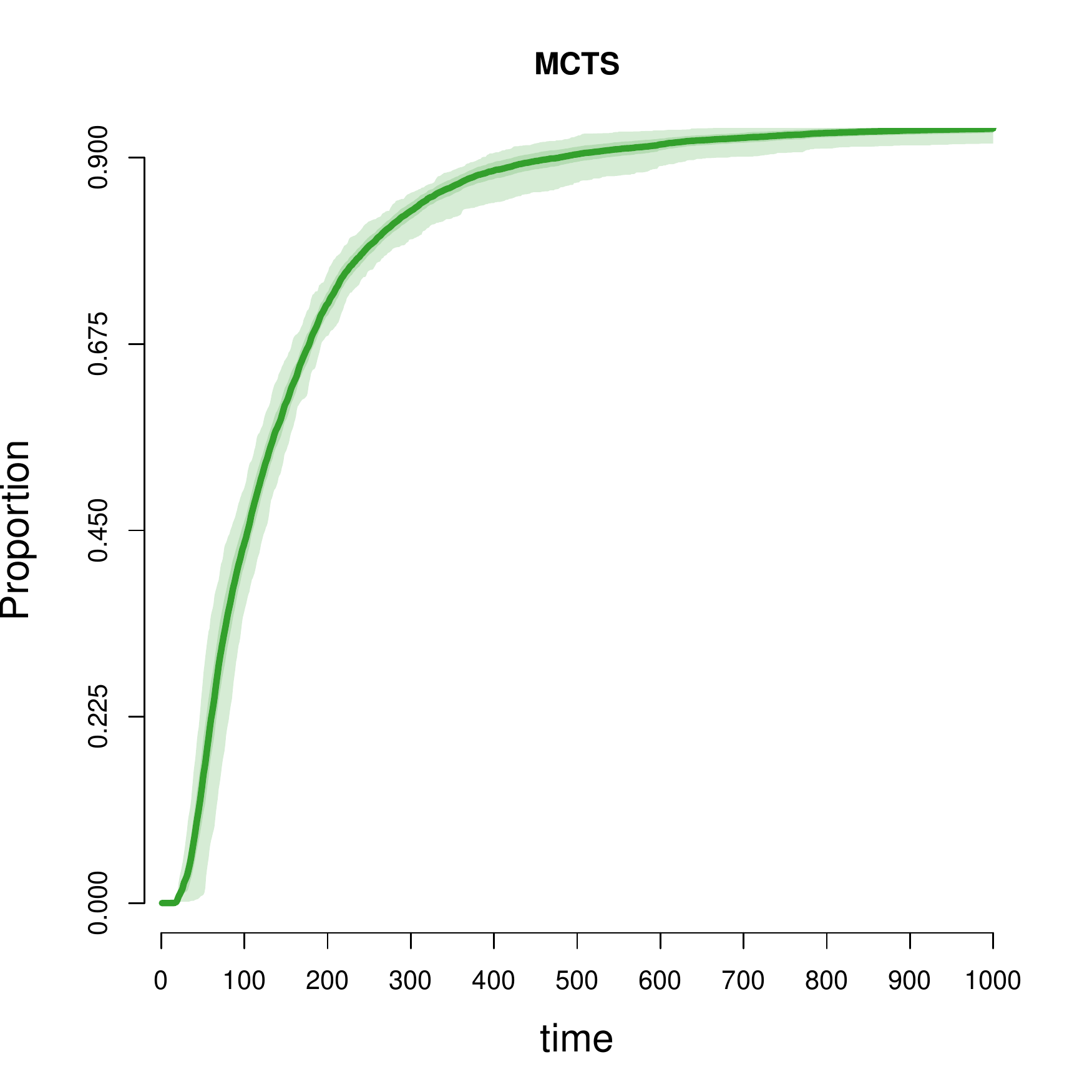}
\includegraphics[width=40mm]{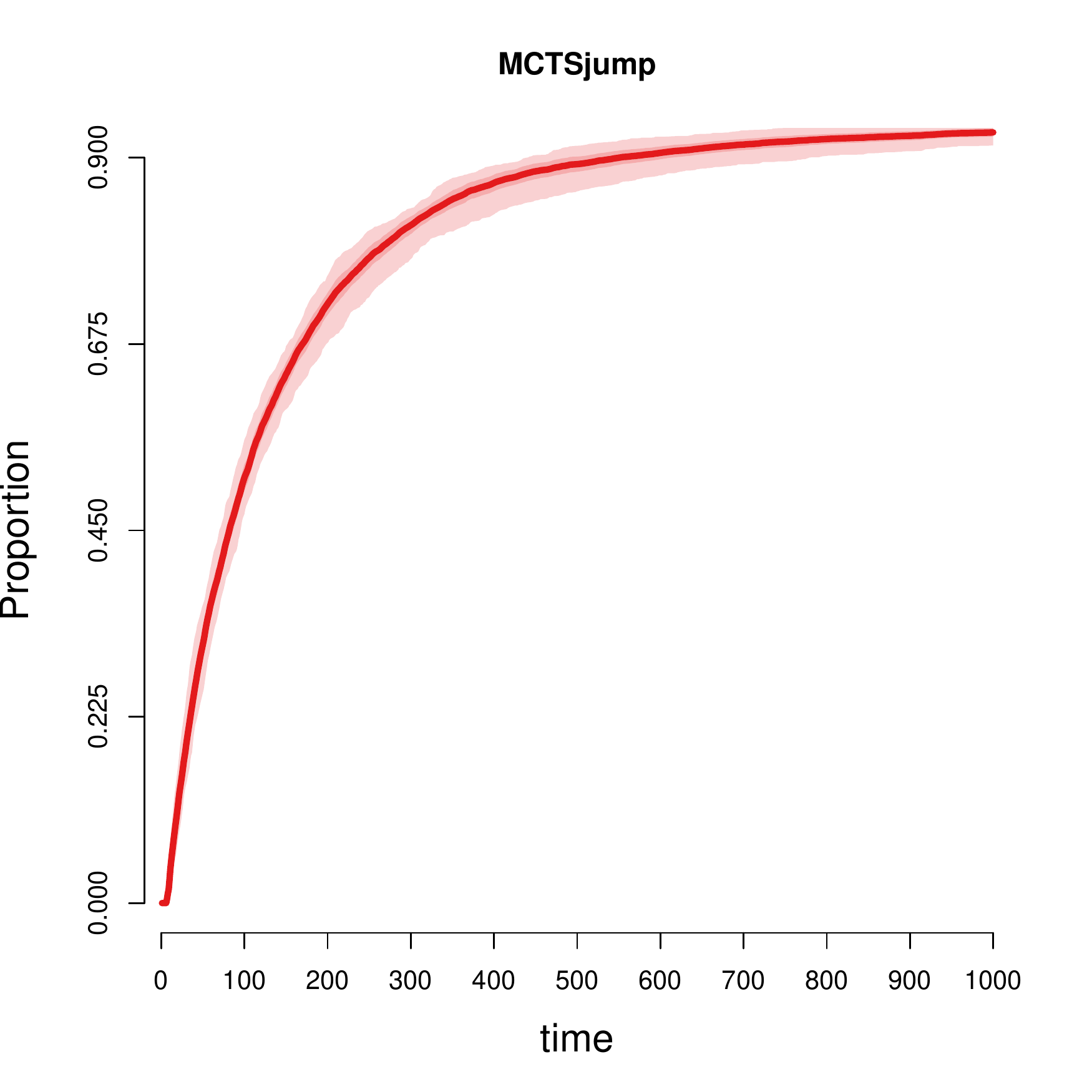} 
\includegraphics[width=40mm]{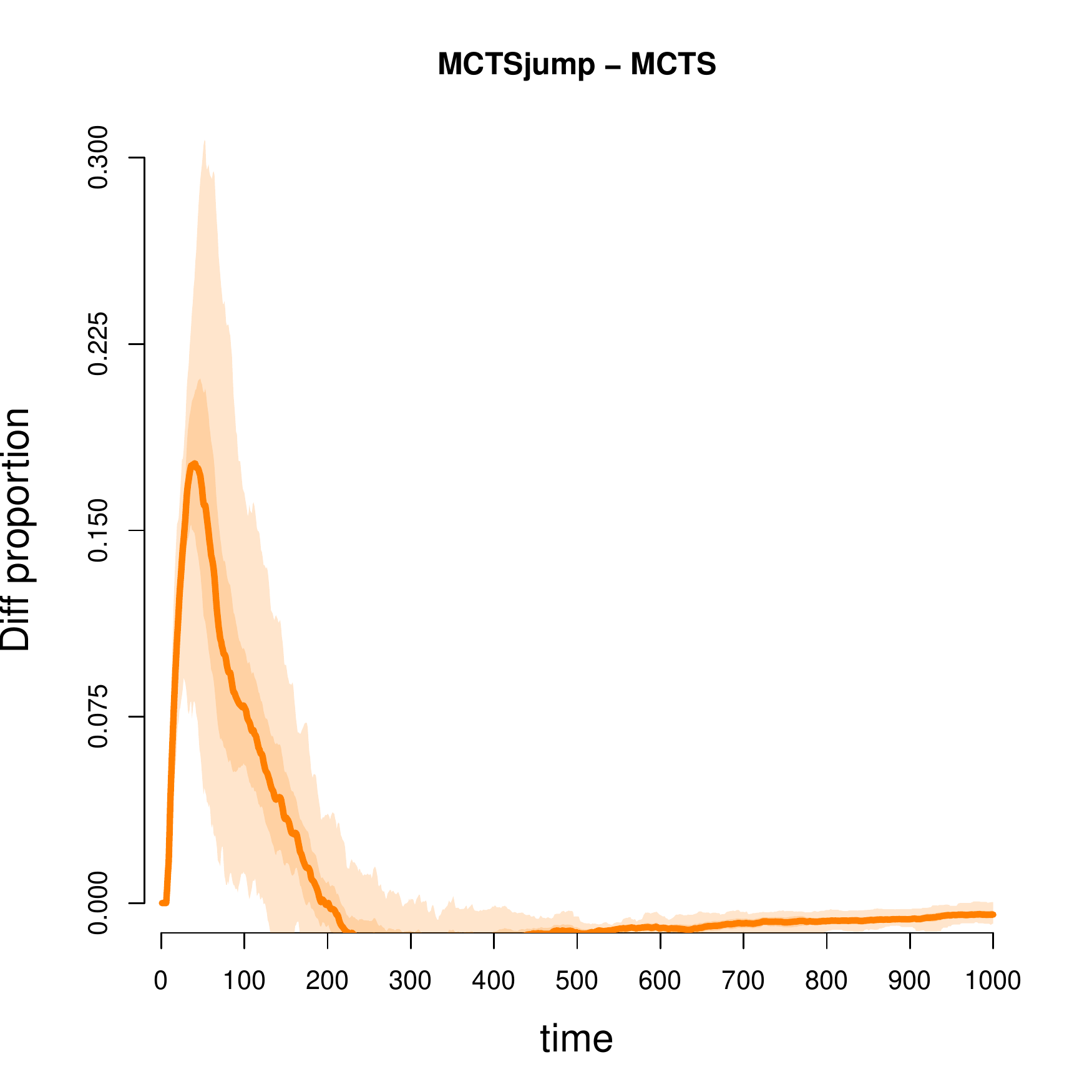}
\includegraphics[width=40mm]{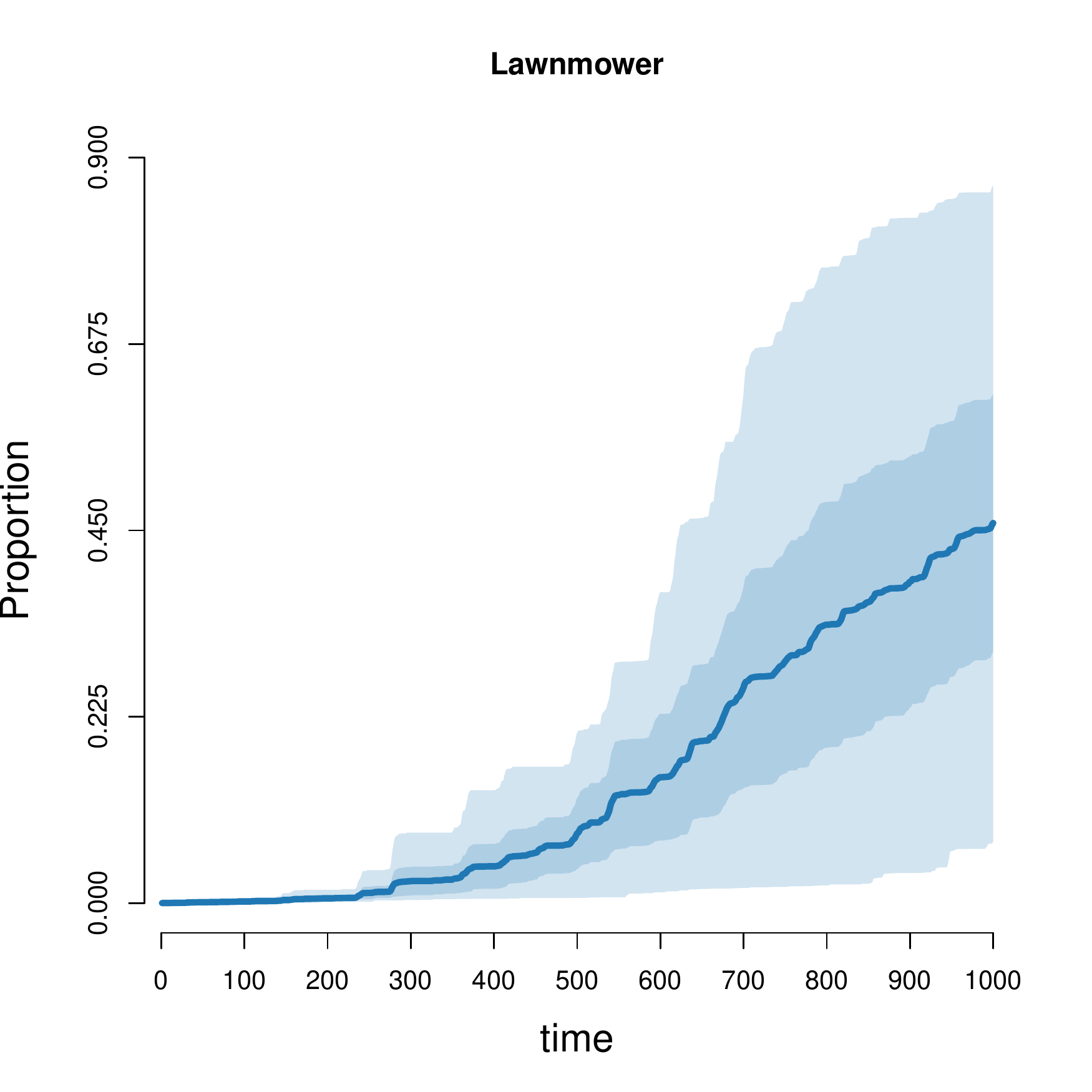}
\includegraphics[width=40mm]{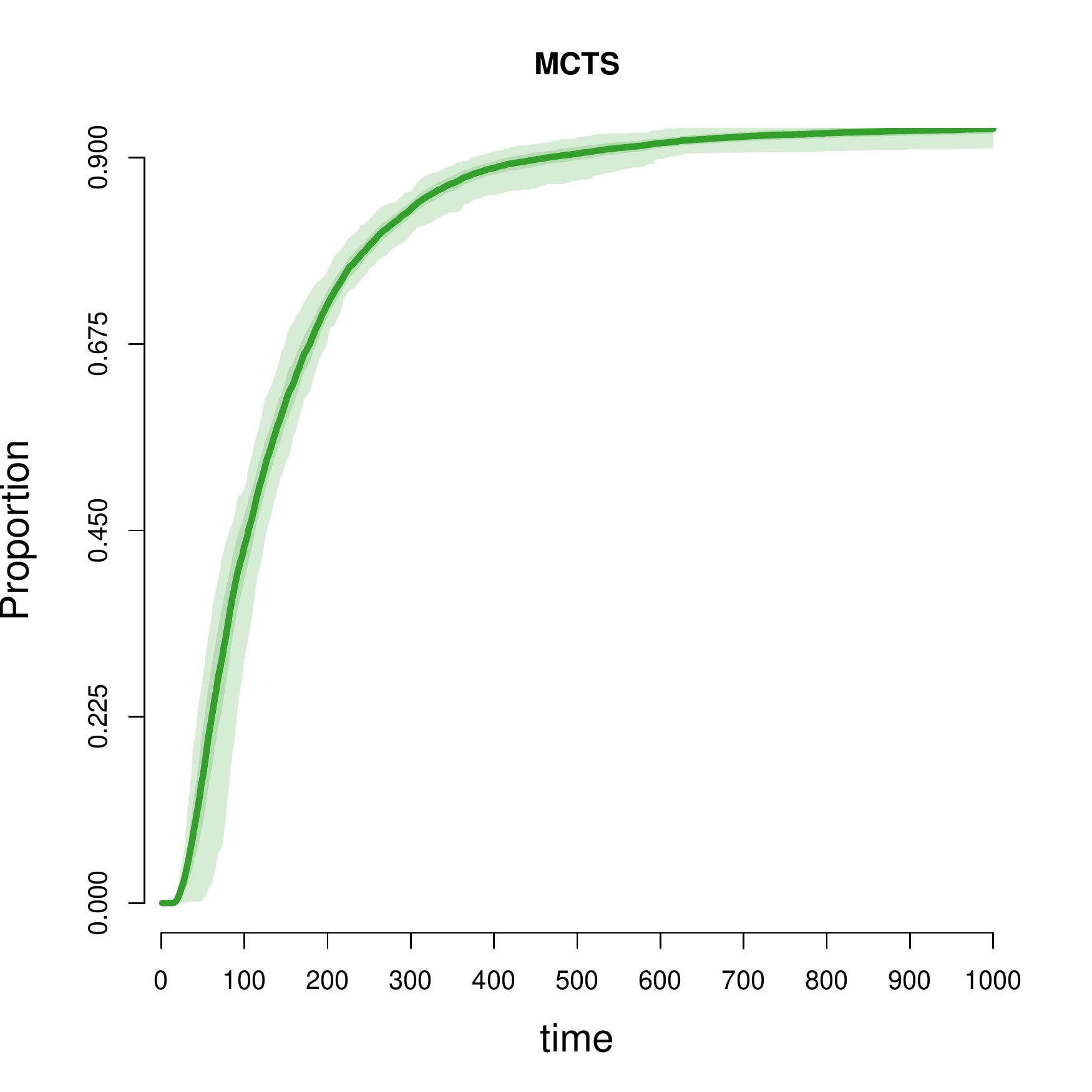}
\includegraphics[width=40mm]{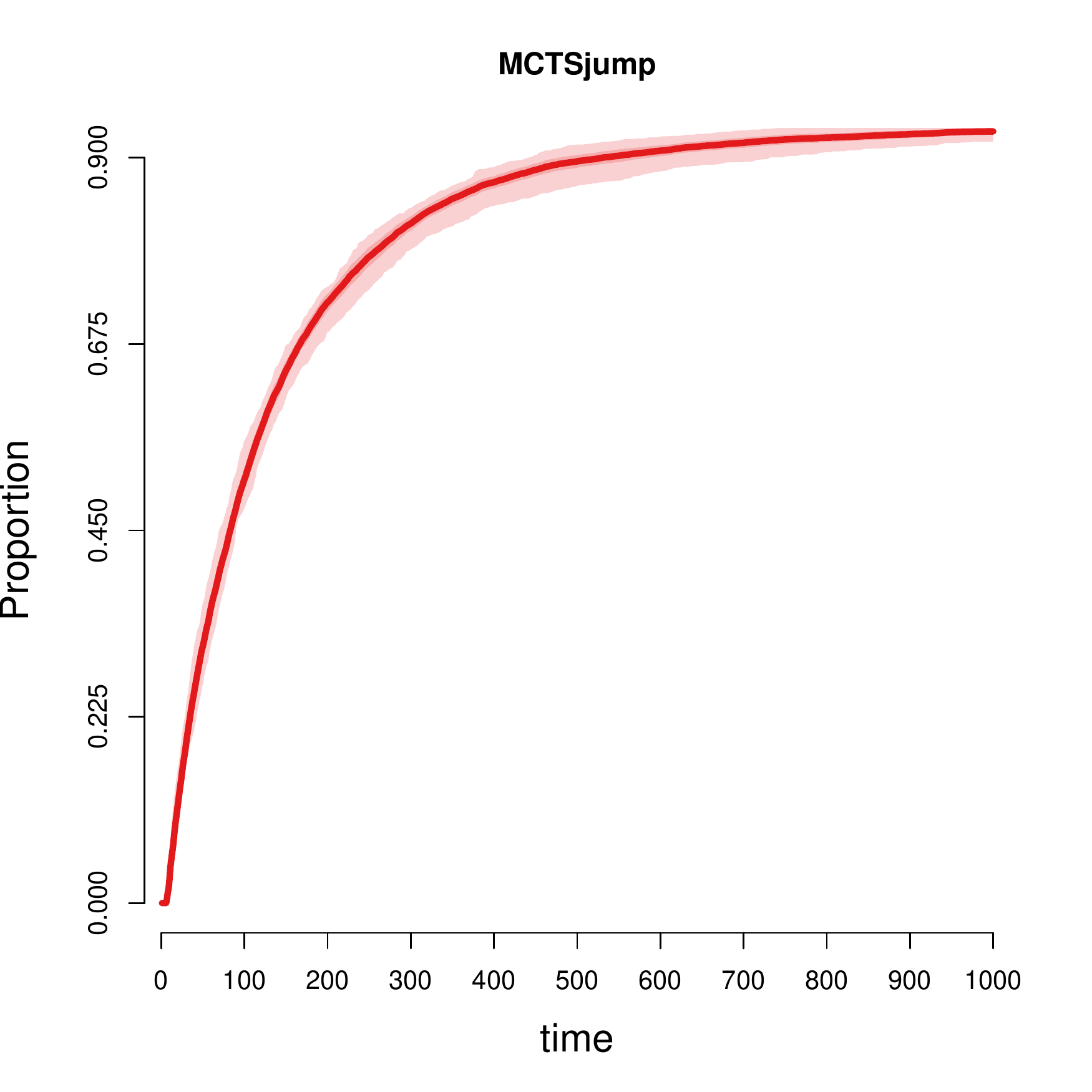} 
\includegraphics[width=40mm]{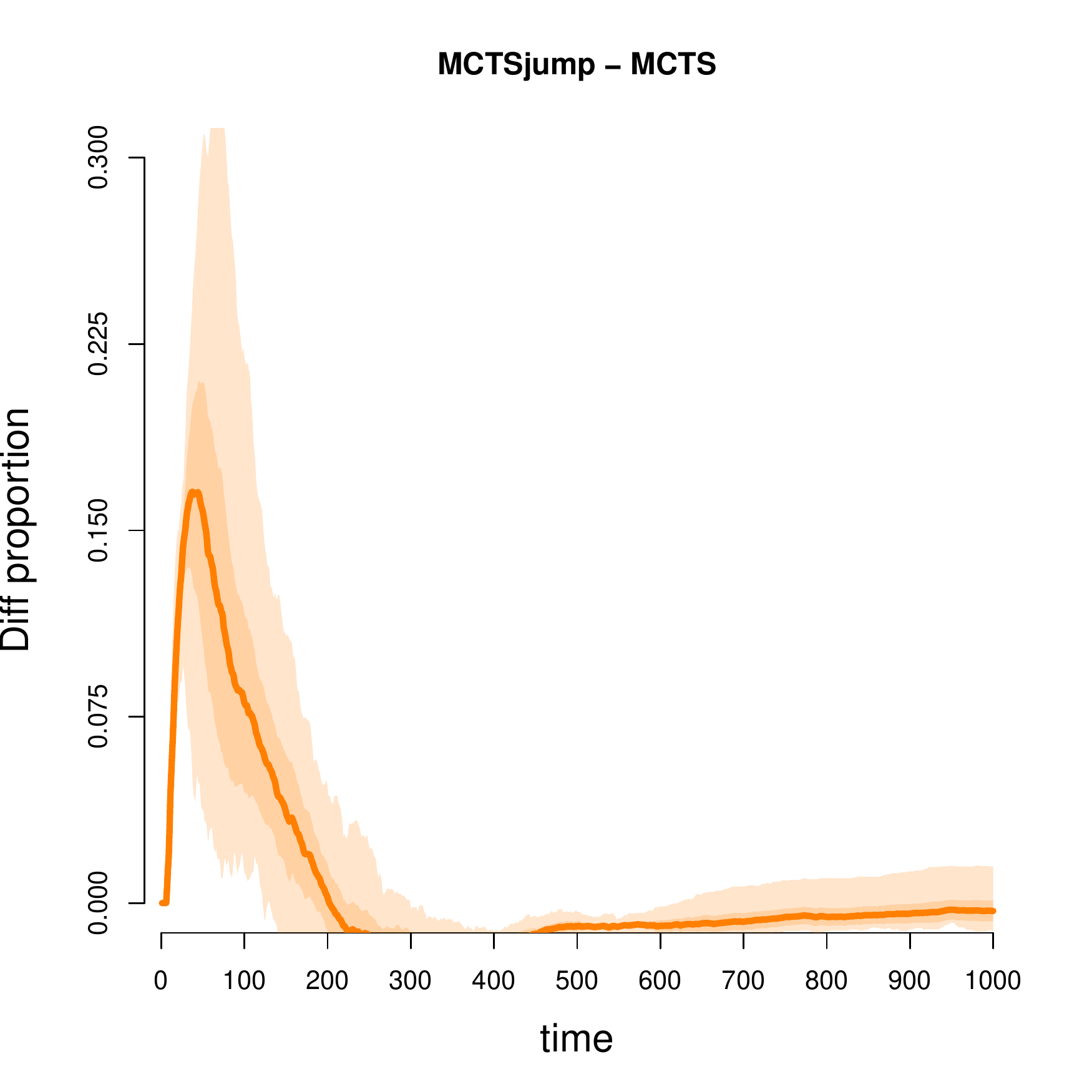}

\caption{Sensitivity analysis for Scenario B. The first row displays the results for the prior used in the paper. The second row changes the prior mean for $\beta$ on buildings and roads in the population intensity to be two standard deviations larger than the baseline prior. The third row changes the prior mean for $\beta$ on buildings in the injury probability to be two standard deviations larger than the baseline prior.}\label{fig:scenario3prop}
\end{figure*}

\end{document}